\documentclass[10pt,twocolumn,letterpaper]{article}

\usepackage[pagenumbers]{cvpr} 

\newcommand\model{S6MOD} 

\definecolor{cvprblue}{rgb}{0.21,0.49,0.74}
\usepackage[pagebackref,breaklinks,colorlinks,allcolors=cvprblue]{hyperref}

\usepackage{booktabs}
\usepackage{nicematrix}
\usepackage{algorithm}
\usepackage{algorithmic}
\usepackage[frozencache,cachedir=_minted-main]{minted}
\setminted{breaklines}
\usepackage{multirow}

\usepackage{setspace}
\usepackage[accsupp]{axessibility}

\def\paperID{2149} 
\def\confName{CVPR}
\def\confYear{2025}

\usepackage{amsmath,amsfonts,bm}

\def\eqref#1{equation~\ref{#1}}

\def\1{\bm{1}}

\def\rmB{{\mathbf{B}}}
\def\rmC{{\mathbf{C}}}

\def\rmF{{\mathbf{F}}}

\def\rmX{{\mathbf{X}}}

\def\rmZ{{\mathbf{Z}}}

\def\vmu{{\bm{\mu}}}

\DeclareMathAlphabet{\mathsfit}{\encodingdefault}{\sfdefault}{m}{sl}
\SetMathAlphabet{\mathsfit}{bold}{\encodingdefault}{\sfdefault}{bx}{n}

\def\gL{{\mathcal{L}}}

\title{Enhancing Online Continual Learning with Plug-and-Play State Space Model and Class-Conditional Mixture of Discretization}

\author{Sihao Liu$^{1}$ \quad Yibo Yang$^{2}$\footnotemark[2] \quad Xiaojie Li$^{3}$ \quad David A. Clifton$^{4}$ \quad Bernard Ghanem$^{2}$ \\
{\normalsize $^1$Harbin Institute Of Technology} \quad
{\normalsize $^2$King Abdullah University of Science and Technology} \quad \\
{\normalsize $^3$Harbin Institute Of Technology (Shenzhen)} \quad 
{\normalsize $^4$University of Oxford} \quad \\
{\normalsize liusihaowo@gmail.com \quad yibo.yang93@gmail.com \quad xiaojieli0903@gmail.com \quad} \\
{\normalsize david.clifton@eng.ox.ac.uk \quad bernard.ghanem@kaust.edu.sa}\\
}

\begin{document}
\maketitle
\begin{abstract}
\renewcommand{\thefootnote}{\fnsymbol{footnote}}

Online continual learning (OCL) seeks to learn new tasks from data streams that appear only once, while retaining knowledge of previously learned tasks. 
Most existing methods rely on replay, focusing on enhancing memory retention through regularization or distillation. However, they often overlook the adaptability of the model, limiting the ability to learn generalizable and discriminative features incrementally from online training data.
To address this, we introduce a plug-and-play module, \model, which can be integrated into most existing methods and directly improve adaptability. Specifically, {\model} introduces an extra branch after the backbone, where a mixture of discretization selectively adjusts parameters in a selective state space model, enriching selective scan patterns such that the model can adaptively select the most sensitive discretization method for current dynamics.
We further design a class-conditional routing algorithm for dynamic, uncertainty-based adjustment and implement a contrastive discretization loss to optimize it. Extensive experiments combining our module with various models demonstrate that {\model} significantly enhances model adaptability, leading to substantial performance gains and achieving the state-of-the-art results.

\footnotetext[2]{Corresponding author.}
\end{abstract}    
\section{Introduction}
\label{sec:intro}

Continual learning (CL) \cite{wang2024comprehensive, parisi2019continual, de2021continual, MAI202228} is a task that requires models to continuously and efficiently learn new ability when receiving new data. 
It tests the model's ability to adapt to dynamically changing environments while retaining existing knowledge. 
For example, autonomous vehicles are expected to continuously learn new driving environments and traffic regulations, thereby improving detection accuracy under complex driving scenes. 
From the perspective of task format, CL can be divided into offline and online \cite{de2021continual, wang2024comprehensive}. 
Unlike offline CL \cite{chaudhry2018riemannian, rebuffi2017icarl}, online CL (OCL) \cite{de2021continual, gunasekara2023survey} 
aligns with real-world implementations as it
requires data to arrive sequentially in mini-batches and allows only one epoch of training, posing greater challenges for efficient adaptation with data accessible for learning only once.



To tackle the challenging OCL, existing studies widely rely on the replay technique \cite{guo2022online} that selectively stores a subset of old-class data to strengthen the memory ability of existing knowledge and mitigate catastrophic forgetting. 
Building on replay, various methods have been proposed with regularization, buffer allocation, and distillation strategies \cite{guo2022online, rolnick2019experience, guo2023dealing, wei2023online, wang2024improving, yan2024orchestrate}. 
For example, MIR \cite{aljundi2019online} aims to mitigate mutual interference among tasks through a retrieval strategy, 
OCM \cite{guo2022online} seeks to reduce forgetting by maximizing mutual information between different tasks, 
and CCLDC \cite{wang2024improving} adopts collaborative learning and distillation to improve plasticity.
However, learning generalizable and discriminative features incrementally from online training data is still intractable, and imprecise features will impede and even mislead the replay strategies \cite{OnPro}. Therefore, inducing accurate and efficient adaptation with limited data access is critically important for OCL and there still remains significant potential for improvement.

Recently, selective state space models (SSMs), also known as S6 in Mamba \cite{gu2023mamba}, have shown promising results in modeling long-range dependencies with computational efficiency. Selective SSMs introduce a selective scan mechanism to make the interactions among sequential states aware of input context, and have been applied to vision tasks by integrating various scan directions \cite{zhu2024vision, Liu2024VMambaVS}. In addition to the success in sequence modeling \cite{lieber2024jamba, das2023decoder, xing2024segmamba, ruan2024vm}, selective SSMs also exhibit stronger adaptability than static parameters in few-shot class-incremental learning due to the dynamic operation weights, as reported in a recent pioneering work \cite{li2024mamba}. 
Despite the increased capacity of adaptation,
directly applying selective SSMs to OCL will be infeasible because it is challenging for selective SSMs to capture a precise 
discretization
pattern with limited context from online data that appears only once for training. Moreover, a single discretization mechanism may fail to capture all the essential details in some nonlinear dynamic systems or systems with multi-scale characteristics \cite{guckenheimer2013nonlinear,abdulle2012heterogeneous}.


To this end, we develop a plug-and-play branch based on selective state space model and class-conditional mixture of discretization. The branch with an equiangular tight frame classifier \cite{yang2023neural, NEURIPS2022_f7f5f501, Zhong_2023_CVPR, XIE202360, MAI202228, seo2024learning} can be integrated on any existing OCL methods to supervise the original classification head and improve the adaptation ability in OCL. 
Considering multiple discretization methods can 
exhibit varying sensitivity to the dynamic characteristics of a system, 
we introduce a mixture of discretization into SSMs to enrich selective scan patterns such that 
the model can adaptively select the most sensitive discretization method for current dynamics. This flexibility allows our method to comprehensively capture the system's complex dynamic features, particularly in cases of rapid state changes and evolving systems as exemplified by OCL. 



In order to guide the mixture of discretization for OCL, where a delicate trade-off between maintaining the stability of old knowledge and fostering the plasticity for new ability is needed, we further introduce a class-conditional routing to aggregate the discretization patterns. Concretely, we maintain a feature prototype for each class to calculate the class uncertainty based on the margin among different classes. For classes with less uncertainty, fewer discretization patterns will be aggregated to stabilize the abilities already acquired for these classes, while for classes with large uncertainty, more discretization patterns will be included to allocate additional capacity for adapting to these undeveloped abilities. After aggregation of the mixture of discretization, we employ a contrastive discretization loss that enforces within-class consistency and between-class diversity, contributing to the learning of generalizable and discriminative features after the selective scan. In experiments, we integrate our plug-and-play branch on numerous OCL methods and significant improvements can be consistently observed on multiple datasets.

 


Our contributions can be summarized as follows:
\begin{itemize}
    \item To enhance the adaptability of existing methods for OCL, we propose \model, a plug-and-play module based on selective state space model and class-conditional mixture of discretization, which can strengthen the base method.   
    \item We further develop a class-conditional gating strategy that dynamically satisfy both objectives of maintaining the stability of knowledge already acquired and fostering the plasticity for undeveloped abilities. A contrastive discretization loss is employed to facilitate the learning of generalizable and discriminative features. 
    \item In experiments, our method can be easily integrated on different OCL methods and is compatible with distillation techniques, to consistently achieve significant improvements on multiple datasets including CIFAR-10, CIFAR-100, and Tiny-ImageNet.
\end{itemize}

\section{Related Work}
\label{sec:formatting}

\paragraph{Continual Learning(CL).}
CL can be categorized into three types based on the implementation methods \cite{de2021continual, wang2024comprehensive, parisi2019continual, MAI202228}: regularization-based, parameter isolation-based, and replay-based. Regularization-based methods \cite{kirkpatrick2017overcoming, zenke2017continual, aljundi2018memory, li2017learning, wang2021afec} introduce regularization terms to restrict changes in model parameters, thereby protecting the already learned knowledge. 
Parameter isolation-based methods \cite{yoon2017lifelong, rusu2016progressive, serra2018overcoming, wortsman2020supermasks, NEURIPS2023_d7b3cef7} typically assign different model parameters or sub-networks to different tasks to avoid interference between tasks. 
Replay-based methods \cite{rebuffi2017icarl, lopez2017gradient, rolnick2019experience, buzzega2020dark, chaudhry2018efficient} perform joint training by either storing a portion of old data or generating pseudo-data. This allows the model to revisit old data while learning new data, thereby reducing forgetting.

\paragraph{Online Continual Learning(OCL).}
OCL is a specialized form of CL designed to test the ability to continuously learn and update as data arrives in real-time streams \cite{de2021continual, gunasekara2023survey}. This means that OCL can only train for a single epoch and typically process only individual samples or mini batches at a time. Replay-based methods are widely used in OCL \cite{guo2022online}. ER \cite{rolnick2019experience} introduces the combination of cross-entropy loss with a random buffer. OCM \cite{guo2022online} uses mutual information maximization to reduce feature bias and preserve past knowledge. GSA \cite{guo2023dealing} proposes a gradient-based adaptive optimization method to address dynamic training bias. OnPro \cite{wei2023online} uses online prototype equilibrium to address shortcut learning problem. Some methods introduce knowledge distillation based on replay. CCL-DC \cite{wang2024improving} introduces Collaborative Continual Learning (CCL) and Distillation Chain (DC) to enhance model plasticity. MOSE \cite{yan2024orchestrate} alleviates forgetting by integrating multi-level supervision and reverse self-distillation. 
However, these models carry a risk when the features used to distill are not precise. 
In comparison, our method {\model}\ as a plug-and-play module is applicable to most OCL methods to improve the adaptability of these models, enabling them to learn generalizable and discriminative features more efficiently.


\paragraph{Selective State Space Model (S6)}
Selective State Space Model (S6) \cite{gu2023mamba} has gained increasing interest
as an alternative to self-attention
\cite{vaswani2017attention} with lower computational complexity. S6 enhances the S4 model \cite{gu2021efficiently} by introducing the selective scan mechanism, and its effectiveness in vision tasks has been extensively studied \cite{zhu2024vision, Liu2024VMambaVS}. 
For example, Vmamba \cite{Liu2024VMambaVS} introduces SS2D, a cross-scanning mechanism for images, facilitating the extension of Mamba to process vision data.
A recent study Mamba-FSCIL \cite{li2024mamba} leverages the dynamic weights and sequence modeling capability of Mamba to achieve dynamic adaptation in few-shot class-incremental learning. 
But their method rely on an effective selective scan learned from training data. In OCL, the online data that appears only once for training provides limited context for Mamba models, and thus poses challenges to capture a precise discretization pattern. Different from these studies, our method integrates S6 with class-conditional mixture of discretization and effectively helps to improve the adaptability for OCL. 
Some existing works have introduced mixture of experts (MoE) into Mamba \cite{pioro2024moe, anthony2024blackmamba, lieber2024jamba}, following the design of switch-transformer \cite{fedus2022switch}. Our method differs from them in that each expert in our mixture of discretization is only a simple linear layer for computation efficiency, and discretization patterns are aggregated with our class-conditional gating to balance between maintaining stability of old knowledge and allocating capacity for learning new abilities. 



\begin{figure*}[t]
  \centering
   \includegraphics[width=1\linewidth]{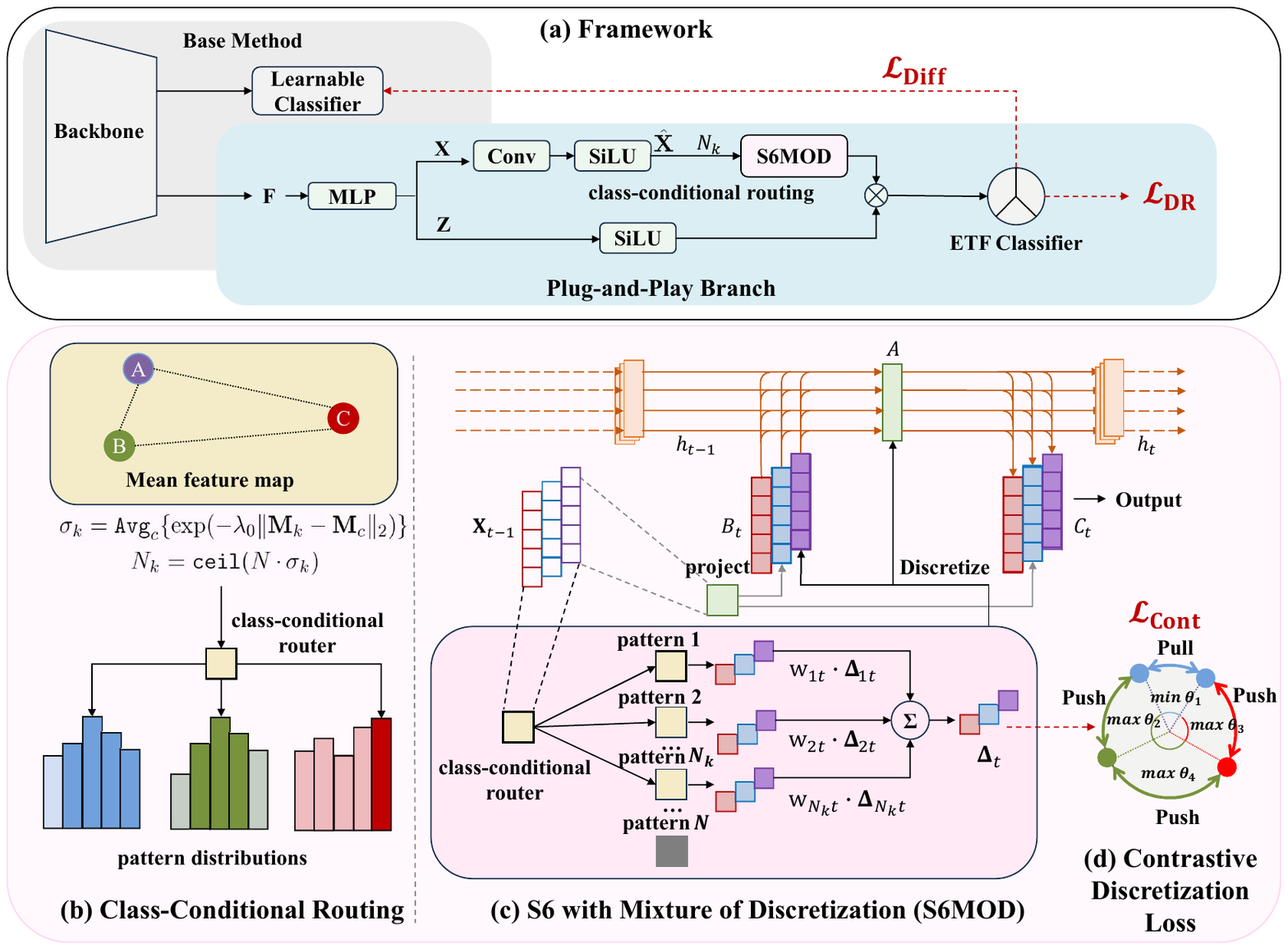}
   \vspace{-22pt}
   \caption{Framework of S6MOD. Our method (a) introduces a plug-and-play branch after the backbone, where features are learned through S6MOD and supervised by the ETF classifier to guide the base method. S6MOD (c) utilizes MoE to enhance the discretization of SSM and applies class-conditional routing (b) to dynamically adjust the discretization based on the uncertainty. Finally, we use a contrastive discretization loss (d) to supervise the learning of both generalizable and discriminative features.}
   \label{fig:framework}
   \vspace{-8pt}
\end{figure*}

\section{Preliminaries}
\subsection{Problem Definition of OCL}
OCL requires a model to continuously update itself from a data stream, with each mini-batch containing new data sampled from a changing distribution \( D_t \). At each time step \( t \), the model receives a mini-batch of data \( \{ (x_i^{(t)}, y_i^{(t)}) \}_{i=1}^{n_t} \), where \( x_i^{(t)} \) represents the input data, \( y_i^{(t)} \) represents the corresponding labels, and \( n_t \) is the number of samples. The goal of OCL is to sequentially update the model to adapt to both gradual and task-specific shifts in the data distribution. 
Given a pre-trained network \( f_\theta \) with parameters \( \theta \), the model updates its parameters by solving an optimization problem \( \arg \min_\theta L(f_\theta(x), y) \), where \( L \) is the loss function, thereby enabling it to adapt effectively to the evolving data distribution.

\subsection{State Space Models}
State Space Models (SSMs) can be viewed as linear time-invariant systems. They map input sequences $x(t) \in \mathbb{R}$ to output sequences $y(t) \in \mathbb{R}$ through hidden states $h(t)\in\mathbb{R}^N$. Mathematically, these models can be represented as linear ordinary differential equations (ODE) :
\begin{equation}
\begin{aligned}
\label{eq:vmu}
h'(t) &= A h(t) + B x(t), \\
y(t) &= C h(t),
\end{aligned}
\end{equation}
where $A\in\mathbb{R}^{N\times N}$ represents the state transition matrix, while $B\in\mathbb{R}^{N\times 1}$ and $C\in\mathbb{R}^{1\times N}$ denote the mapping matrices from input to latent state and from latent state to output, respectively, where $N$ indicates the size of the hidden state.

To apply the SSM to real discrete data, zero-order hold (ZOH) \cite{gu2021efficiently} is employed. It discretizes the continuous parameters $A$,
$B$, and $C$ using the time scale parameter $\bm{\Delta}$. The discretized SSM equations can be rewritten as follows:
\begin{equation}
\begin{aligned}
h_t&=\overline{A}h_{t-1}+\overline{B}x_t,\\
y_t&=Ch_t.
\end{aligned}
\end{equation}

Recently, Gu \cite{gu2023mamba} proposed a new parameterization method for SSM with a selective scan mechanism, which is known as S6 and serves as the core of the Mamba model. To enhance S6's capability in processing visual data, Liu et al.~\cite{Liu2024VMambaVS} introduced SS2D, which can serialize image data from four different directions. Given the input data $x$, the output $\overline{x}$ processed by SS2D can be expressed as follows:
\begin{equation}
\begin{aligned}
\overline{x}=\mathrm{SS2D}(x)=\sum_{i=1}^{4}\mathrm{S6}(\mathrm{scan}(x,i)).
\end{aligned}
\end{equation}

\section{Proposed Method}
In this section, we will first introduce the overall structure of our plug-and-play module {\model}\ in Sec. \ref{framework}. After that, we will present the detailed design of the state space model with mixture of discretization in Sec. \ref{mods6}. Then, we will introduce the class-conditional routing and its corresponding contrastive discretization loss in details in Sec. \ref{routing}. Finally, we will specify the optimization details in Sec. \ref{sumloss}.

\subsection{Overall structure}
\label{framework}

To better enhance existing methods' ability in learning generalizable and discriminative features incrementally from online training data, 
we propose a plug-and-play module named {\model}\ that can be easily applied to existing OCL methods, as shown in Fig. \ref{fig:framework} (a).
Overall, {\model}\ strengthens the original method by introducing a plug-and-play branch after the backbone. It consists of a block and a fixed equiangular tight frame (ETF) classifier \cite{NEURIPS2022_f7f5f501, yang2023neural}. The intermediate features $\rmF$ generated by the backbone are duplicated and sent into the classification head of the original base method, and the selective state space model (SSM) branch introduced by our method.


In our SSM branch, $\rmF$ is projected into two paths through an MLP. The first one $\rmX = f_x({\rmF})$ is used to perform selective scan with our class-conditional mixture of discretization, as will be detailed in later subsections. The other one $\rmZ = f_z({\rmF})$ performs a gating mechanism as commonly adopted in Mamba models \cite{mehta2022long, Liu2024VMambaVS, li2024mamba}.
The output feature of this branch can be formulated as:
\begin{equation}
\begin{aligned}
\label{eq:vmu}
\vmu = \text{SiLU}(\rmZ) \otimes \text{S6MOD}(\rm{SiLU} (\text{Conv}(\rmX))).
\end{aligned}
\end{equation}

To assist the learning of the base method for generalizable and discriminative features, we adopt a regularization, $\gL_\text{Diff}$, on the predicted distribution of the base method, which is the KL divergence between $P$ and $Q$ and can be formulated as:
\begin{equation}
\label{eq:diff}
\gL_\text{Diff}=\sum_iP(i)\log\left(\frac{P(i)}{Q(i)}\right),
\end{equation}
where $P$ represents the predicted distribution of the base method, and $Q$ refers to the output prediction after the ETF classifier of our SSM branch, \emph{i.e.,} $Q=\rm{softmax}(\mathbf{W}_{\rm{ETF}}\vmu)$.


\subsection{S6 with Mixture of Discretization}
\label{mods6}
We integrate the selective space state model with a mixture of discretization, borrowing the concept of mixture of experts \cite{6215056, shazeer2017outrageously, 6797059, eigen2013learning}. 
Its structure is shown in Fig. \ref{fig:framework} (c). Similar to the typical SSM structure, we also use an MLP layer to produce the projection matrices $B$ and $C$ from the input features, which can be expressed as follows:
\begin{equation}
\begin{aligned}
\label{equ:params}
\rmB = f_B(\hat{\rmX}), \quad 
\rmC = f_C(\hat{\rmX}),
\end{aligned}
\end{equation}
where $\hat{\rmX}$ denotes the input of the selective scan module S6MOD, \emph{i.e.,} $\hat{\rmX}=\rm{SiLU}(\rm{Conv}(\rmX))$. 
To enable selective SSM to capture a precise discretization pattern with limited context from online data, 
we develop a sparse MoE system for the discretization transformation $\bm{\Delta}$ in S6 models, 
utilizing specialized projection layers to enrich the discretization patterns and adaptively selecting the most sensitive discretization based on the current dynamics.
Specifically, each discretization candidate is produced by a projection layer as follows:
\begin{equation}
\begin{aligned}
\label{eq:expert}
\bm{\Delta}_i = f_{\Delta i}(\hat{\rmX}).
\end{aligned}
\end{equation}
The input feature $\rmX$ is also fed into a sparse gating mechanism, which produces associated importance weights $w_i$ corresponding to each discretization $\bm{\Delta}_i$. Through the class-conditional gating that will be introduced in Sec. \ref{routing}, we dynamically control the number of discretization patterns, $N_k$, to be selected for each class $k$ based on the input features $\hat{\rmX}$. Finally, the selected discretization patterns are aggregated with their importance weights as follows:
\begin{equation}
\label{eq:weiexpert}
\bm{\Delta} = \sum_{i \in \Omega(\hat{\rmX})} w_i\cdot \bm{\Delta}_i,\quad |\Omega(\hat{\rmX})|=N_k,
\end{equation}
where $\Omega(\hat{\rmX})$ is the index set of the top-$N_k$ discretization candidates of $\hat{\rmX}$, according to their importance weights.




It's noteworthy that $\bm{\Delta}$ is important because it controls the decay rate during the state update process, 
enabling the model to flexibly select and retain important information when handling long sequences, while avoiding the accumulation of redundant and irrelevant information \cite{gu2023mamba}, especially in cases of rapid changing states and evolving systems, as exemplified by OCL. Our method with mixture of discretization facilitates the learning of a precise selective scan pattern with limited data context in OCL.

\subsection{Class-Conditional Routing}
\label{routing}

With more discretization patterns selected, the model is able to allocate more capacity for learning new knowledge, but may cause a significant shift of the model that impairs the existing ability. If we can calculate $N_k$ based on the mis-classification probability of each class $k$, we can dynamically control the capacity of mixture of discretization based on the learning conditions of all classes.  
Therefore, in order to guide the mixture of discretization to strike a balance between maintaining the stability of old knowledge and fostering the plasticity for new ability, we further design class-conditional routing.
During training, we maintain the feature prototypes $\mathbf{M}=\{\mathbf{M}_c\}$ by moving average, where $\mathbf{M}_c$ is the within-class feature mean for class $c$. 
Then, we estimate the class uncertainty $\sigma_k$ based on the average margin of class $k$ with different classes,
\begin{equation}
\label{eq:vmu}
\sigma_k = \texttt{Avg}_{c} \{\exp(-\lambda_0 \| \mathbf{M}_k-\mathbf{M}_c \|_2) \},
\end{equation}
where 
$\lambda_0$ is a hyper-parameter
and $\sigma_k$ denotes the class uncertainty for class $k$ ranging from $(0, 1)$. A large $\sigma_k$ indicates that the class center $k$ has narrow margins with the other class centers, and thus tends to be misclassified, while a small $\sigma_k$ happens when class center $k$ is distant from the other class centers with a lower likelihood of being misclassified. 
In inference, we replace $\mathbf{M}_k$ with the input feature $\hat{\mathbf{X}}$ and calculate its uncertainty by Eq. (\ref{eq:vmu}) with the feature prototypes $\mathbf{M}_c$ of all classes. 



Then, we multiply the uncertainty by the number of total discretization candidates $N$, to get the discretization number to select for class $k$ or input feature $\hat{\mathbf{X}}$ as follows,
\begin{equation}
\begin{aligned}
\label{eq:nk}
N_{k}=\texttt{ceil}(N \cdot \sigma_k),   
\end{aligned}
\end{equation}
where $\verb|ceil|$ refers to the operation that rounds up $N \cdot \sigma_k$ to the nearest integer.

The institution is that classes that are more prone to misclassification have smaller margins with the other classes in the feature space, and thus needs to aggregate more discretization patterns to allocate additional capacity for adapting to these undeveloped abilities. 
Conversely, for categories that are easier to classify with low uncertainty, a smaller $N_k$ is favored to only include the most likely discretization patterns, which can spare optimization efforts for the uncertain classes and stabilize these already acquired abilities. 
The combination of class-conditional routing and our mixture of discretization achieves dynamic structures dependent on input features and classes, such that the final prediction of each class selectively optimizes the corresponding discretization patterns. 



To further reduce interaction among different classes and facilitate the learning of generalizable and discriminative features, we also incorporate contrastive discretization loss function $\gL_\text{Cont}$, as shown in Fig. \ref{fig:framework} (d). It encourages $\bm{\Delta}$ to have within-class consistency and between-class diversity as follows:
\begin{equation}
\label{eq:aggloss}
\mathcal{L}_{\text{Cont}} = - \frac{1}{B^2} \sum_{m=1}^B \sum_{n=1}^B \left( 1_{y_m = y_n} - 1_{y_m \neq y_n} \right) \frac{\bm{\Delta}_m \cdot \bm{\Delta}_n}{\|\bm{\Delta}_m\| \|\bm{\Delta}_n\|}
\end{equation}
where $B$ is the batch size, $\bm{\Delta}_m$ represents the aggregated $\bm{\Delta}$ by Eq. (\ref{eq:weiexpert}) for input $m$, $y_m$ denotes the class of $m$ and $1_{y_m = y_n}$ is an indicator function that takes a value of 1 when the condition $y_m = y_n$ holds, and 0 otherwise.


\subsection{Optimization}
\label{sumloss}
In addition to the previously mentioned loss functions in Eqs. (\ref{eq:diff}) and (\ref{eq:aggloss}), we also directly used the classification loss function $\mathcal{L}_{\text{DR}}$ specifically designed for ETF classifier \cite{NEURIPS2022_f7f5f501, yang2023neural} to supervise our introduced plug-and-play branch,
and the loss function $\mathcal{L}_{\text{z}}$ for maintaining load balancing among the discretization patterns \cite{zoph2022st}.
In summary, when integrating {\model}\ on a base method, the overall loss function can be expressed as follows:\
\begin{equation}
\mathcal{L}_{\text{all}} = \mathcal{L}_{\text{base}} + \mathcal{L}_{\text{S6MOD}},
\label{eq:all}
\end{equation}
\begin{equation}
    \mathcal{L}_{\text{S6MOD}} = \mathcal{L}_{\text{DR}}  + \alpha \cdot \mathcal{L}_{\text{Diff}} + \beta \cdot \mathcal{L}_{\text{Cont}} +  \mathcal{L}_{\text{z}},
\end{equation}
where $\alpha$ and $\beta$ are hyperparameters.
\section{Experiments}
\label{experiments}
\subsection{Experimental Setups}
\label{results}
\begin{table*}
    \centering
    \caption{Average Accuracy (\%, higher is better) on three benchmark datasets with difference memory buffer size $M$, with and without our proposed {\model} module. All values are averages of 10 runs.}
    \vspace{-8pt}
    \resizebox{0.88\textwidth}{!}{
    \begin{tabular}{lccccccccc}
\toprule
\multicolumn{1}{c}{Dataset} & \multicolumn{2}{c}{CIFAR10} & \multicolumn{3}{c}{CIFAR100} & \multicolumn{3}{c}{Tiny-ImageNet}\\
\cmidrule(lr){1-1}
\cmidrule(lr){2-3}
\cmidrule(lr){4-6}
\cmidrule(lr){7-9}
\multicolumn{1}{c}{Memory Size $M$} & 500 & 1000 & 1000 & 2000 & 5000 & 2000 & 5000 & 10000\\
\midrule
ER~\cite{rolnick2019experience} & 56.68{\scriptsize ±1.89} & 62.32{\scriptsize ±4.13} & 24.47{\scriptsize ±0.72} & 31.89{\scriptsize ±1.45} & 39.41{\scriptsize ±1.81} & 10.82{\scriptsize ±0.79} & 19.16{\scriptsize ±1.42} & 24.71{\scriptsize ±2.52}\\
ER + Ours & \textbf{57.88{\scriptsize ±3.30}} & \textbf{65.80{\scriptsize ±2.16}} & \textbf{26.50{\scriptsize ±2.23}} & \textbf{34.55{\scriptsize ±1.66}} & \textbf{39.61{\scriptsize ±3.16}} & \textbf{10.94{\scriptsize ±1.47}} & \textbf{19.67{\scriptsize ±1.36}} & \textbf{25.62{\scriptsize ±1.73}}\\
\midrule
OCM~\cite{guo2022online} & 68.19{\scriptsize ±1.75} & 73.15{\scriptsize ±1.05} & 28.02{\scriptsize ±0.74} & 35.69{\scriptsize ±1.36} & 42.22{\scriptsize ±1.06} & 18.36{\scriptsize ±0.95} & 26.74{\scriptsize ±1.02} & 31.94{\scriptsize ±1.19}\\
OCM + Ours & \textbf{70.57{\scriptsize ±1.14}} & \textbf{75.31{\scriptsize ±1.10}} & \textbf{32.44{\scriptsize ±1.35}} & \textbf{38.97{\scriptsize ±2.28}} & \textbf{45.49{\scriptsize ±1.58}} & \textbf{19.12{\scriptsize ±1.60}} & \textbf{27.20{\scriptsize ±1.83}} & \textbf{32.31{\scriptsize ±1.72}}\\
\midrule
OnPro~\cite{wei2023online} & 70.47{\scriptsize ±2.12} & 74.70{\scriptsize ±1.51} & 27.22{\scriptsize ±0.77} & 33.33{\scriptsize ±0.93} & 41.59{\scriptsize ±1.38} & 14.32{\scriptsize ±1.40} & 21.13{\scriptsize ±2.12} & 26.38{\scriptsize ±2.18}\\
OnPro + Ours& \textbf{72.30{\scriptsize ±1.08}} & \textbf{75.14{\scriptsize ±0.99}} & \textbf{28.84{\scriptsize ±0.59}} & \textbf{37.57{\scriptsize ±0.81} }& \textbf{44.14{\scriptsize ±1.36}} & \textbf{17.27{\scriptsize ±0.65}} & \textbf{25.62{\scriptsize ±0.79}} & \textbf{29.20{\scriptsize ±1.00}}\\
\midrule
OCM-CCLDC~\cite{wang2024improving} & 74.14{\scriptsize ±0.85} & 77.66{\scriptsize ±1.46} & 35.00{\scriptsize ±1.15} & 43.34{\scriptsize ±1.51} & 51.43{\scriptsize ±1.37} & 23.36{\scriptsize ±1.18} & 33.17{\scriptsize ±0.97} & 39.25{\scriptsize ±0.88}\\
OCM-CCLDC + Ours& \textbf{74.45{\scriptsize ±1.20}} & \textbf{78.21{\scriptsize ±1.03}} & \textbf{36.02{\scriptsize ±1.77}} & \textbf{44.40{\scriptsize ±2.26} }& \textbf{52.53{\scriptsize ±0.21}} & \textbf{23.68{\scriptsize ±1.02}} & \textbf{33.59{\scriptsize ±0.72}} & \textbf{39.53{\scriptsize ±1.02}}\\
\midrule
OnPro-CCLDC~\cite{wang2024improving} & 74.49{\scriptsize ±2.14} & 78.64{\scriptsize ±1.42} & 34.76{\scriptsize ±1.12} & 41.89{\scriptsize ±0.82} & 50.01{\scriptsize ±0.85} & 21.81{\scriptsize ±1.02} & 32.00{\scriptsize ±0.72} & 38.18{\scriptsize ±1.02}\\
OnPro-CCLDC + Ours& \textbf{75.03{\scriptsize ±1.03}} & \textbf{79.51{\scriptsize ±0.63}} & \textbf{36.07{\scriptsize ±0.76}} & \textbf{43.92{\scriptsize ±0.96} }& \textbf{50.85{\scriptsize ±0.52}} & \textbf{22.03{\scriptsize ±0.82}} & \textbf{33.29{\scriptsize ±0.35}} & \textbf{38.51{\scriptsize ±0.91}}\\
\midrule
MOSE~\cite{yan2024orchestrate} & 61.02{\scriptsize ±1.47} & 70.74{\scriptsize ±1.18} & 35.05{\scriptsize ±0.34} & 45.06{\scriptsize ±0.32} & 54.53{\scriptsize ±0.78} & 18.23{\scriptsize ±0.73} & 30.98{\scriptsize ±0.63} & 38.71{\scriptsize ±0.44}\\
MOSE + Ours& \textbf{63.74{\scriptsize ±1.55}} & \textbf{72.54{\scriptsize ±0.14}} & \textbf{35.43{\scriptsize ±0.62}} & \textbf{45.38{\scriptsize ±0.31}}& \textbf{54.75{\scriptsize ±0.52}} & \textbf{19.03{\scriptsize ±0.83}} & \textbf{32.03{\scriptsize ±0.81}} & 38.18{\scriptsize ±1.02}\\
\midrule
MOE-MOSE~\cite{yan2024orchestrate} & 62.54{\scriptsize ±1.59} & 72.18{\scriptsize ±1.29} & 37.32{\scriptsize ±0.34} & 47.03{\scriptsize ±0.57} & 55.62{\scriptsize ±0.72} & 20.61{\scriptsize ±0.69} & 32.52{\scriptsize ±0.33} & 38.41{\scriptsize ±0.53}\\
MOE-MOSE + Ours& \textbf{63.86{\scriptsize ±1.19}} & \textbf{73.16{\scriptsize ±0.53}} & \textbf{37.76{\scriptsize ±0.51}} & \textbf{47.25{\scriptsize ±0.43}}& \textbf{56.32{\scriptsize ±0.53}} & \textbf{21.03{\scriptsize ±0.77}} & \textbf{33.53{\scriptsize ±0.51}} & \textbf{40.11{\scriptsize ±0.26}}\\
\bottomrule
\end{tabular}
    }

    \label{tab:aa}
    \vspace{-8pt}
\end{table*}
    
\begin{table*}
    \centering
    \caption{Average Forgetting (\%, lower is better) on three benchmark datasets with difference memory buffer size $M$, with and without our proposed {\model} module. All values are averages of 10 runs.}
    \vspace{-8pt}
    \resizebox{0.88\textwidth}{!}{
    \begin{tabular}{lccccccccc}
\toprule
\multicolumn{1}{c}{Dataset} & \multicolumn{2}{c}{CIFAR10} & \multicolumn{3}{c}{CIFAR100} & \multicolumn{3}{c}{Tiny-ImageNet}\\
\cmidrule(lr){1-1}
\cmidrule(lr){2-3}
\cmidrule(lr){4-6}
\cmidrule(lr){7-9}
\multicolumn{1}{c}{Memory Size $M$} & 500 & 1000 & 1000 & 2000 & 5000 & 2000 & 5000 & 10000\\
\midrule
ER~\cite{rolnick2019experience} & 33.16{\scriptsize ±3.50} & 20.94{\scriptsize ±6.79} & 32.65{\scriptsize ±1.78} & 22.20{\scriptsize ±2.26} & 13.29{\scriptsize ±1.98} & 58.38{\scriptsize ±1.69} & 46.87{\scriptsize ±1.60} & 40.77{\scriptsize ±2.45}\\
ER + Ours & \textbf{31.25{\scriptsize ±3.44}} & \textbf{16.71{\scriptsize ±2.86}} & \textbf{30.96{\scriptsize ±2.07}} & \textbf{19.02{\scriptsize ±2.13}} & \textbf{12.97{\scriptsize ±2.61}} & \textbf{58.03{\scriptsize ±2.19}} & \textbf{45.83{\scriptsize ±1.49}} & \textbf{37.62{\scriptsize ±1.53}}\\
\midrule
OCM~\cite{guo2022online} & 13.68{\scriptsize ±4.25} & 11.63{\scriptsize ±2.62} & 14.99{\scriptsize ±1.55} & 9.16{\scriptsize ±1.75} & 3.76{\scriptsize ±1.16} & 26.12{\scriptsize ±1.63} & 19.74{\scriptsize ±1.30} & 15.92{\scriptsize ±1.47}\\
OCM + Ours & \textbf{13.58{\scriptsize ±3.26}} & \textbf{9.68{\scriptsize ±3.47}} & 15.99{\scriptsize ±1.82} & 11.60{\scriptsize ±1.49} & 5.46{\scriptsize ±0.98} & \textbf{24.23{\scriptsize ±2.07}} & \textbf{19.12{\scriptsize ±1.87}} & \textbf{14.66{\scriptsize ±1.75}}\\
\midrule
OnPro~\cite{wei2023online} & 17.94{\scriptsize ±3.69} & 14.20{\scriptsize ±2.60} & 16.76{\scriptsize ±2.47} & 12.42{\scriptsize ±1.39} & 6.72{\scriptsize ±0.94} & 28.01{\scriptsize ±1.59} & 23.52{\scriptsize ±1.75} & 20.32{\scriptsize ±1.70}\\
OnPro + Ours& \textbf{7.13{\scriptsize ±1.44}} & \textbf{5.73{\scriptsize ±1.81}} & \textbf{16.03{\scriptsize ±1.69}} & \textbf{9.81{\scriptsize ±1.32} }& \textbf{5.07{\scriptsize ±0.89}} & \textbf{20.92{\scriptsize ±1.12}} & \textbf{16.82{\scriptsize ±0.97}} & \textbf{18.15{\scriptsize ±1.91}}\\
\midrule
OCM-CCLDC~\cite{wang2024improving} & 11.59{\scriptsize ±2.24} & 9.18{\scriptsize ±2.03} & 16.69{\scriptsize ±2.36} & 10.07{\scriptsize ±1.37} & 3.99{\scriptsize ±0.78} & 26.16{\scriptsize ±1.90} & 19.99{\scriptsize ±1.96} & 15.56{\scriptsize ±1.06}\\
OCM-CCLDC + Ours& 15.81{\scriptsize ±2.69} & 11.28{\scriptsize ±1.83} & \textbf{16.64{\scriptsize ±1.85}} & \textbf{9.08{\scriptsize ±2.06} }& \textbf{3.48{\scriptsize ±0.31}} & \textbf{25.29{\scriptsize ±1.95}} & \textbf{14.89{\scriptsize ±0.74}} & \textbf{11.55{\scriptsize ±1.31}}\\
\midrule
OnPro-CCLDC~\cite{wang2024improving} & 19.89{\scriptsize ±4.01} & 14.62{\scriptsize ±2.75} & 28.93{\scriptsize ±2.19} & 20.23{\scriptsize ±1.03} & 10.55{\scriptsize ±1.89} & 28.21{\scriptsize ±1.58} & 20.85{\scriptsize ±1.13} & 16.17{\scriptsize ±0.63}\\
OnPro-CCLDC + Ours& \textbf{17.23{\scriptsize ±3.16}} & \textbf{11.51{\scriptsize ±2.33}} & \textbf{22.26{\scriptsize ±1.18}} & \textbf{12.49{\scriptsize ±1.59} }& \textbf{4.88{\scriptsize ±1.36}} & \textbf{26.73{\scriptsize ±1.51}} & \textbf{16.69{\scriptsize ±0.31}} & \textbf{12.23{\scriptsize ±1.37}}\\
\midrule
MOSE~\cite{yan2024orchestrate} & 30.36{\scriptsize ±1.69} & 20.27{\scriptsize ±1.27} & 37.54{\scriptsize ±0.43} & 25.89{\scriptsize ±0.45} & 13.60{\scriptsize ±0.59} & 47.16{\scriptsize ±1.41} & 24.96{\scriptsize ±0.62} & 15.51{\scriptsize ±0.33}\\
MOSE + Ours& \textbf{27.38{\scriptsize ±1.94}} & \textbf{17.92{\scriptsize ±0.34}} & \textbf{37.09{\scriptsize ±0.63} } & \textbf{25.80{\scriptsize ±0.39}}& 15.55{\scriptsize ±0.61} & \textbf{46.60{\scriptsize ±1.07}} & \textbf{24.39{\scriptsize ±0.72}} & 20.15{\scriptsize ±0.74}\\
\midrule
MOE-MOSE~\cite{yan2024orchestrate} & 29.39{\scriptsize ±1.79} & 19.24{\scriptsize ±1.49} & 35.17{\scriptsize ±0.30} & 23.99{\scriptsize ±0.51} & 12.81{\scriptsize ±0.74} & 41.98{\scriptsize ±1.46} & 22.22{\scriptsize ±0.38} & 13.94{\scriptsize ±0.59}\\
MOE-MOSE + Ours& \textbf{28.13{\scriptsize ±1.41}} & \textbf{17.98{\scriptsize ±0.56}} & \textbf{34.76{\scriptsize ±0.66}} & \textbf{23.84{\scriptsize ±0.77}}& 14.03{\scriptsize ±0.48} & \textbf{41.26{\scriptsize ±1.02}} & \textbf{21.39{\scriptsize ±0.58}} & 14.91{\scriptsize ±0.28}\\
\bottomrule
\end{tabular}
    }
    \label{tab:la}
    \vspace{-1.4em}
\end{table*}


\paragraph{Datasets.} We test three datasets that are widely used in OCL, i.e., CIFAR-10 (10 classes) \cite{Krizhevsky2009LearningML}, CIFAR-100 (100 classes) \cite{Krizhevsky2009LearningML} and TinyImageNet (200 classes) \cite{Le2015TinyIV}. The dataset settings we adopted are the same as those in CCLDC\cite{wei2023online}. We divide CIFAR-10 into 5 tasks, with 2 classes per task; CIFAR-100 into 10 tasks, with 10 classes per task; and TinyImageNet into 100 tasks, with 2 classes per task. Further details about the datasets will be provided in the supplementary materials.

\paragraph{Baselines.} To demonstrate the effectiveness and applicability of our module, we conduct tests on 7 typical and state-of-the-art methods, including ER \cite{rolnick2019experience}, OCM \cite{guo2022online}, OnPro \cite{wei2023online}, OCM-CCLDC \cite{wang2024improving}, OnPro-CCLDC \cite{wang2024improving}, MOSE\cite{yan2024orchestrate} and MOE-MOSE \cite{yan2024orchestrate}.

\paragraph{Implementation details.} To ensure a fair comparison, we applied the same hyperparameter settings to each baseline and the methods combined with our module ({\model}). All the above methods use ResNet-18 as the backbone, without pretraining. For the streaming input data, we set the batch size to 10, and for the samples drawn from the buffer, the batch size is set to 64. We retain each baseline's original data augmentation methods without modifying the base methods. For more details, please refer to the supplementary materials.

\subsection{Results}
We combine our method with both classical and state-of-the-art approaches on CIFAR-10, CIFAR-100, and Tiny-ImageNet. The experimental results in Table \ref{tab:aa} demonstrate the universality of our method, as it consistently improves accuracy across numerous baselines. On CIFAR-10 and CIFAR-100, our method generally leads to a 1\% improvement, while on Tiny-ImageNet, we achieve significant gains under OCM, OnPro and MOE-MOSE. Notably, in settings like CIFAR100 ($M=2k$) and Tiny-ImageNet ($M=5k$), OnPro result in an approximate 4.2\% and 4.5\% improvement, respectively. It is also worth mentioning that even for state-of-the-art distillation-based methods like CCLDC and MOE-MOSE, the performance improvements are quite substantial. For instance, OnPro-CCLDC with the {\model} integration achieve an impressive 79.51\% on CIFAR10 ($M=1k$), and MOE-MOSE with {\model} reach 56.32\% on CIFAR100 ($M=5k$) and 40.1\% on Tiny-ImageNet ($M=10k$). 

In addition, as shown in Table \ref{tab:la}, our module is also highly effective in reducing model forgetting, achieving lower forgetting rates in most settings, with some cases showing particularly significant improvements. This can be attributed to our class-conditional routing. In a few baseline settings, the inclusion of our module led to a higher forgetting rate, but this does not imply that the model is more prone to forgetting. In these cases, we think the higher forgetting rates are often due to the stronger learning capabilities of the model. Analyzing both Table \ref{tab:aa} and Table \ref{tab:la}, we can find that even in these settings with higher forgetting rates, the accuracy performance remains quite competitive.

\subsection{Ablation Studies}
\begin{table}
    \centering
    \caption{Ablation studies on CIFAR-100 ($M=2k$). The " branch" refers to using SS2D\cite{Liu2024VMambaVS} as the core of the {\model}, without routing and $\mathcal{L}_{\text{Cont}}$. "routing" represents class-conditional routing. S6MOD refers to the composition of (branch + class-conditional routing + $\mathcal{L}_{\text{Cont}}$). All values are averaged over 10 runs. }
    \vspace{-5pt}
    \resizebox{0.8\linewidth}{!}{
    \begin{tabular}{lcc}
\toprule
\multicolumn{1}{c}{Method} & Acc. ↑ & AF ↓ \\
\midrule
OnPro & 33.33{\scriptsize ±0.93} & 12.42{\scriptsize ±1.39}\\
OnPro + branch (with $\gL_\text{Diff}$) & 36.61{\scriptsize ±1.04} & 10.28{\scriptsize ±1.59}\\
OnPro + branch + routing & 36.92{\scriptsize ±0.72} & 8.85{\scriptsize ±0.81}\\
OnPro + branch + $\mathcal{L}_{\text{Cont}}$ & 36.93{\scriptsize ±0.84} & 9.70{\scriptsize ±1.42}\\
OnPro + \model & 37.57{\scriptsize ±0.81} & 9.81{\scriptsize ±1.32}\\
\bottomrule
\end{tabular}
    }
    \label{tab:ablation}
    \vspace{-1.6em}
\end{table}

As mentioned in Sec.~\ref{sec:intro}, the {\model} module we propose is very simple and can be directly applied to existing baselines. By adding an extra branch and using the $\mathcal{L}_{\text{Diff}}$ to supervise the original classification head, we can achieve a significant improvement. To demonstrate this, we conduct an ablation study where we only add the extra branch and $\mathcal{L}_{\text{Diff}}$ to the baseline. The results in Table \ref{tab:ablation} confirm this, showing a 3.3\% improvement with supervision from the extra branch alone.

In addition, to verify the effectiveness and adaptability of class-conditional routing and contrastive discretization loss $\mathcal{L}_{\text{Cont}}$, we conduct separate ablation experiments. As shown in Table \ref{tab:ablation}, when added to the base method with the extra branch, the improvements brought by adding class-conditional routing or $\mathcal{L}_{\text{Cont}}$ alone are not significant, with accuracy gains of only 0.31\% and 0.32\%, respectively. However, when both components are used together, the improvement becomes significant, with an approximate 1\% increase in accuracy, and the forgetting rate still decreases. This indicates that when our module is correctly combined, it can significantly enhances the model's adaptability by learning more generalizable and discriminative features through dynamic parameters.

\subsection{Analysis}

\begin{figure}
\begin{center}
    \subfloat[OnPro]{
       \includegraphics[width=0.40\linewidth]{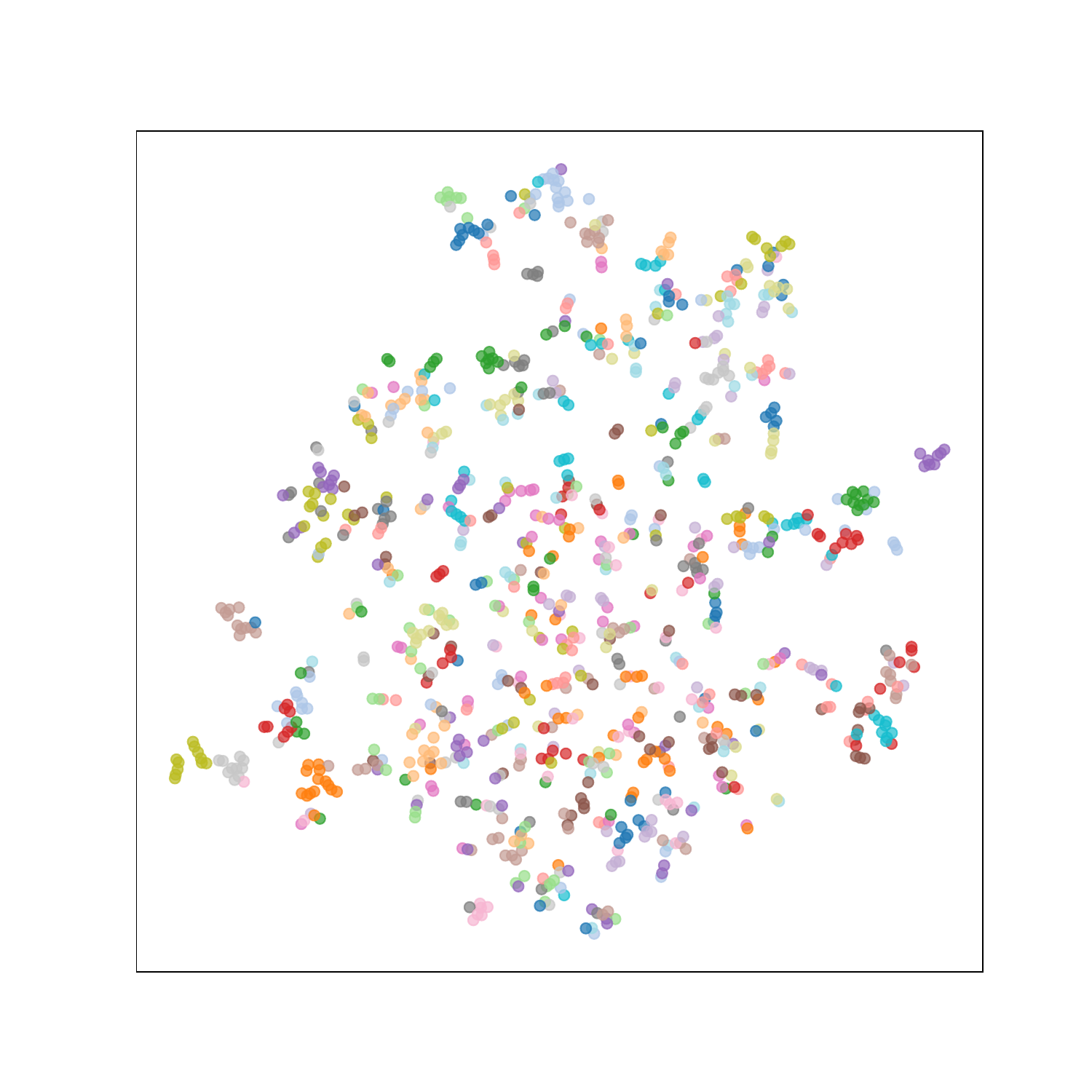}}
    \label{1a}
    \subfloat[OnPro + {\model}]{
        \includegraphics[width=0.40\linewidth]{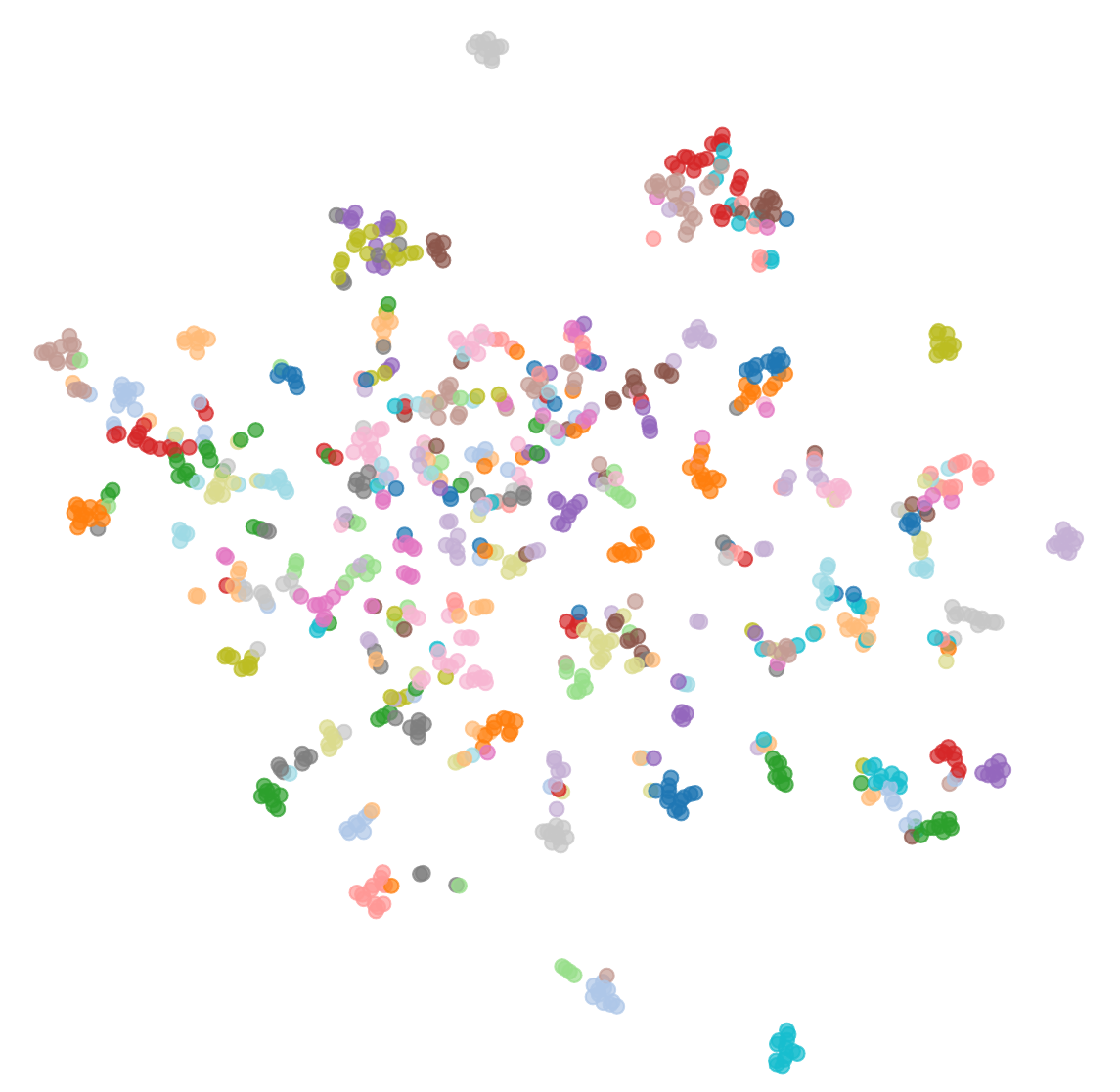}}
    \label{1b}
    \subfloat[OCM]{
        \includegraphics[width=0.40\linewidth]{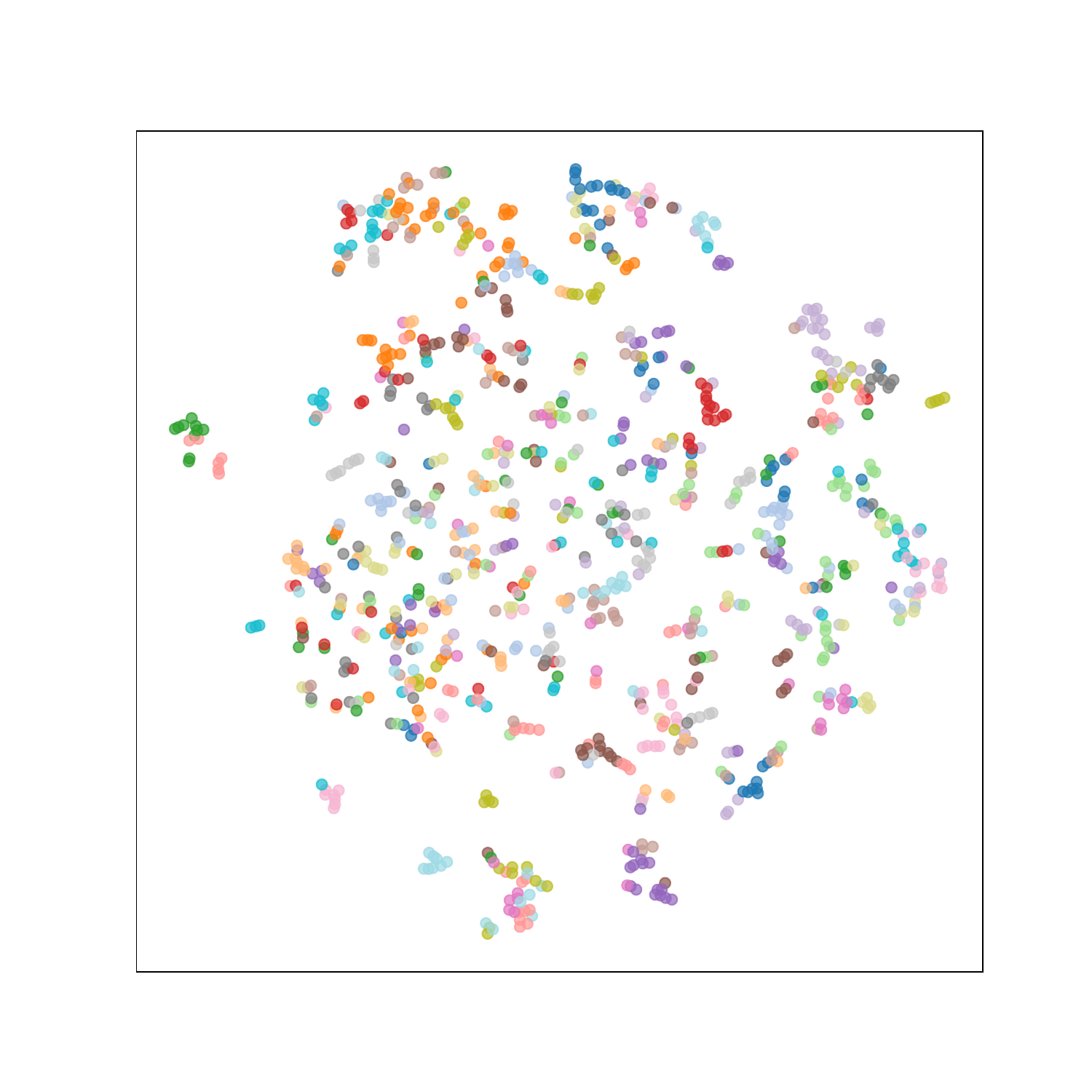}}
    \label{1b}
    \subfloat[OCM + {\model}]{
        \includegraphics[width=0.40\linewidth]{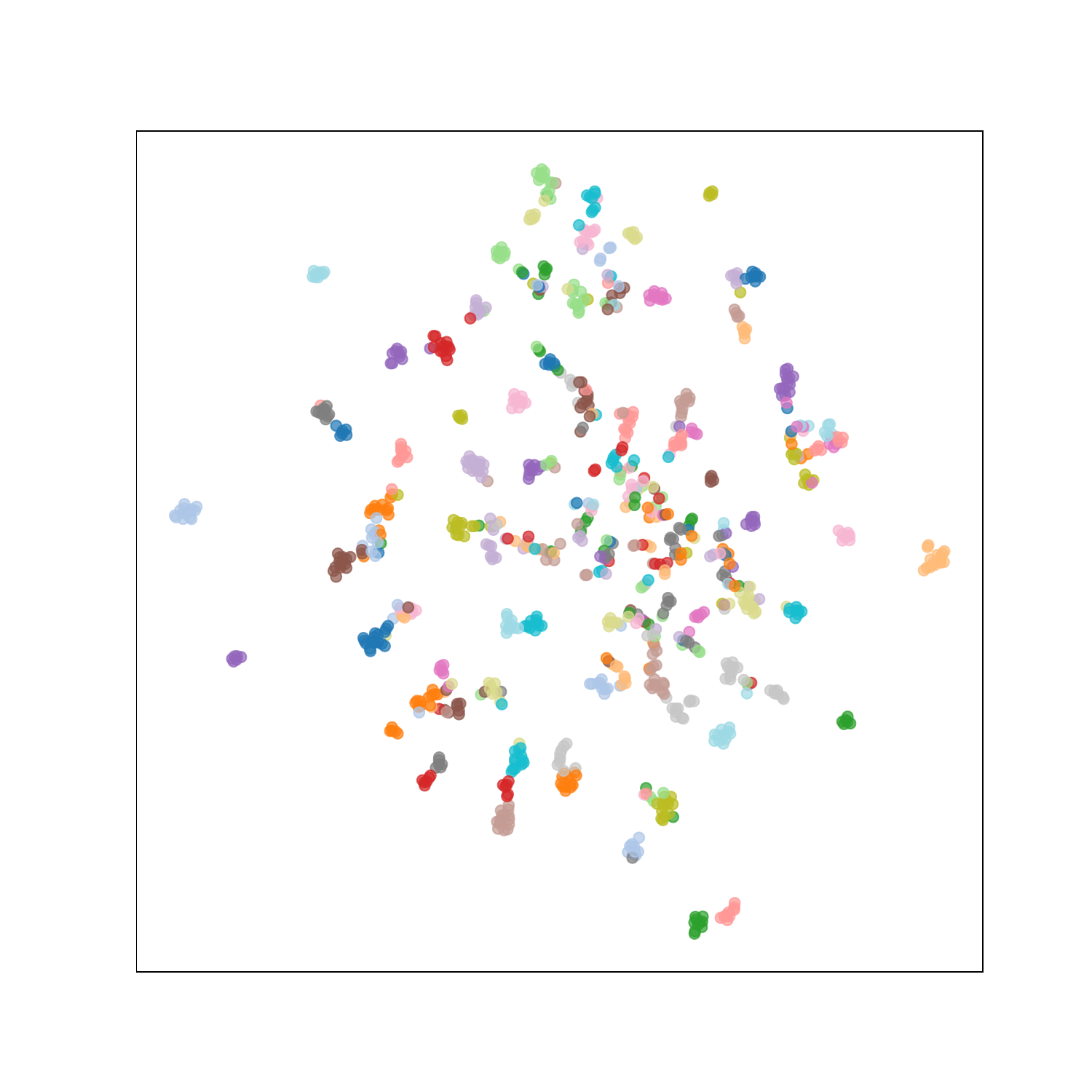}}
    \label{1b}
    \caption{t-SNE visualization of memory data at the end of training on CIFAR-100 ($M=2k$), showcasing baseline in (a) and (c) and baseline combined with {\model} in (b) and (d). Different colors represent different classes.}
    \label{fig:randaug} 
    \vspace{-26pt}
\end{center}
\end{figure}

\paragraph{Analysis of Feature Embeddings.} To verify that {\model} improves the adaptability of the baseline and helps it learn more discriminative features, we use t-SNE \cite{JMLR:v9:vandermaaten08a} visualization for analysis. We compare with the baseline under the CIFAR-100 ($M=2k$) setting. As shown in Fig.~\ref{fig:randaug}, we visualize the features of the samples in the buffer after the final training, which are the features $\rmF$ output by the backbone just before they are passed to the classifier for classification. The features shown in Fig.~\ref{fig:randaug} (a) and (c) are very scattered, with only a few classes showing some degree of aggregation. Most classes do not have clear separations between other classes, making it difficult to distinguish. In contrast, Fig.~\ref{fig:randaug} (b) and (d) present the feature distribution after adding our module. We can observe that many categories have noticeably converged, forming small clusters with consistent intra-class compactness and clear inter-class separations. These features are much easier to distinguish and more conducive to classification by the classifier, which aligns with the significant improvements we demonstrate in the experiments.

\begin{figure}
\begin{center}
    \subfloat[New-Task Accuracy]{
       \includegraphics[width=0.488\linewidth]{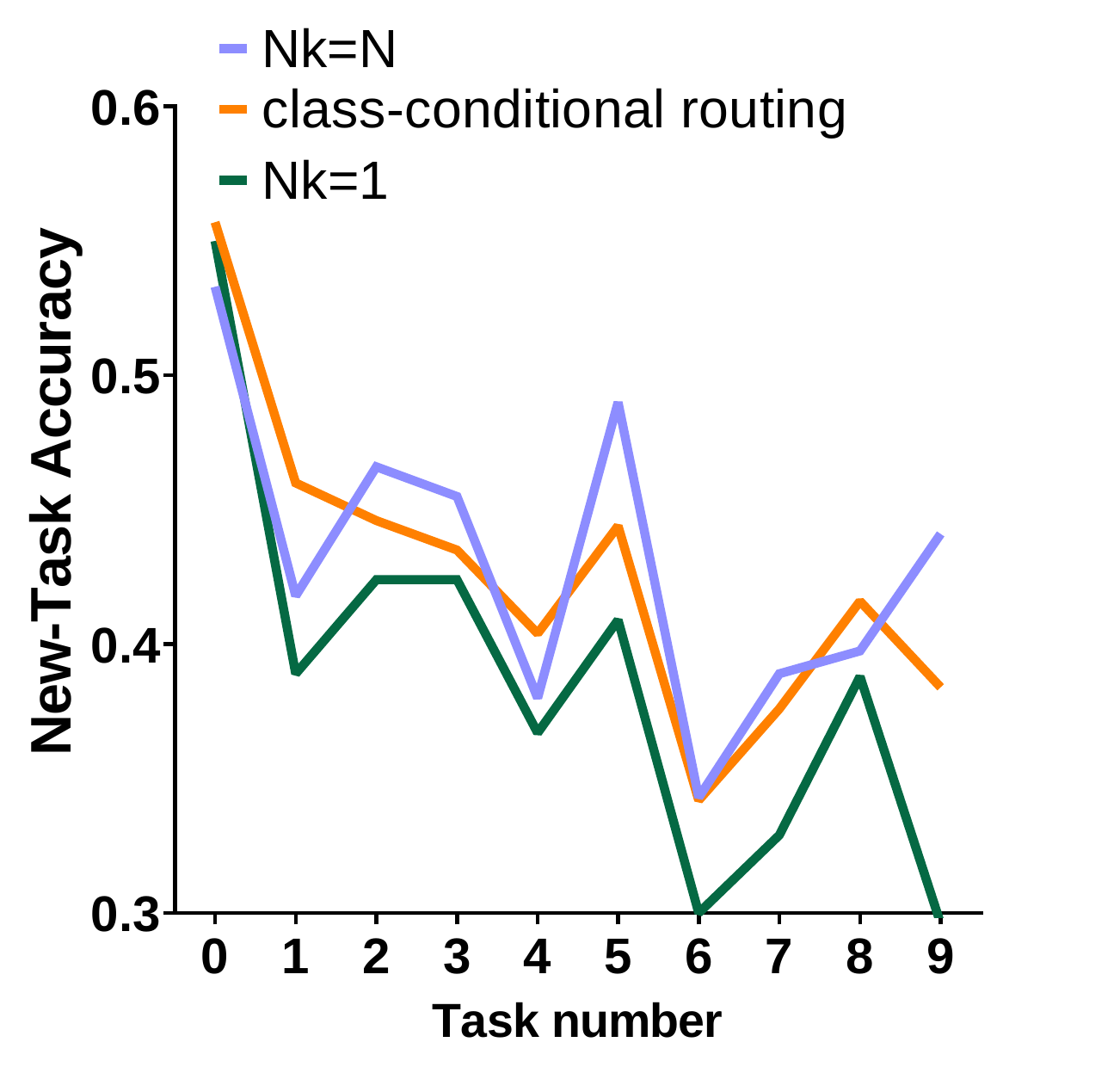}}
    \label{1a}\hfill
    \subfloat[Average Forgetting]{
        \includegraphics[width=0.49\linewidth]{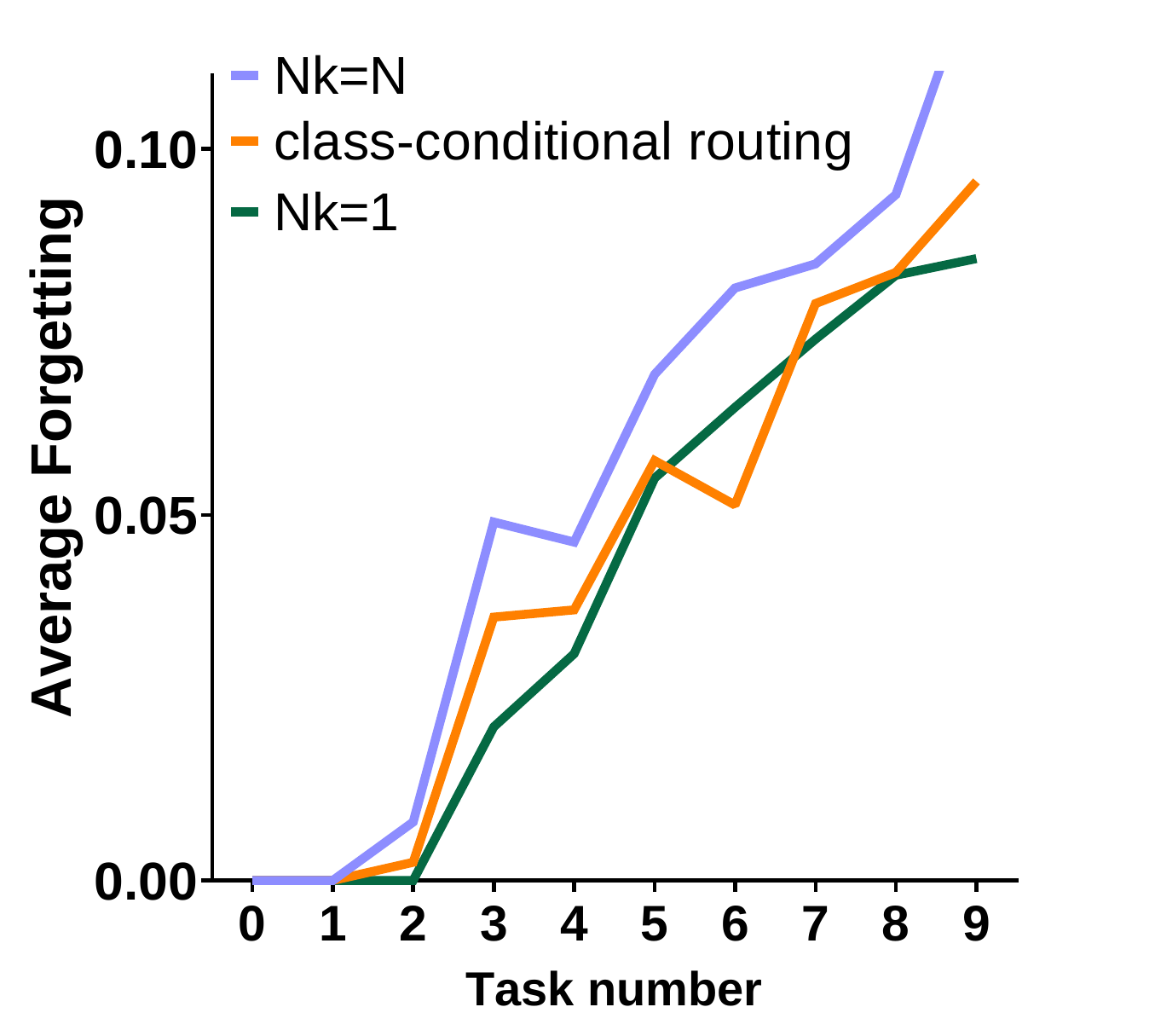}}
    \label{1b}
    \vspace{-8pt}
    \caption{Impact of dynamically selecting different $N_k$ values on the ability to learn new tasks and prevent forgetting of old tasks: New-Task accuracy represents the model's accuracy on the current task. We conduct experiments by setting $N_k = 1$, $N_k = N$, and calculating $N_k$ through class-conditional routing. The dataset used is CIFAR-100 ($M=2k$).}
    \label{fig:topk} 
    \vspace{-24pt}
\end{center}
\end{figure}

\vspace{-4pt}
\paragraph{Analysis of Class-conditional Routing.} We further investigate the effectiveness of class-conditional routing. Under the CIFAR-100 ($M=2k$) setting, we design experiments to compare the impact of a fixed $N_k$ and dynamically calculated $N_k$ using class-conditional routing. Specifically, we set $N_k=N$, meaning all patterns are activated, and $N_k=1$, meaning only the pattern with the highest probability is activated. To facilitate analysis, we present the accuracy of the model on the latest task (the current task) in Fig.~\ref{fig:topk} (a). As shown in Fig.~\ref{fig:topk} (a), when $N_k=N$, all patterns are activated, and thanks to the inclusion of more patterns with different weights and larger parameters, the model has greater capacity to adapt to new tasks, resulting in a stronger ability to learn new knowledge. When $N_k=1$, only the pattern with the highest probability is activated, significantly reducing the model's ability to learn the new task. However, as shown in Fig.~\ref{fig:topk} (b), when $N_k=N$, since all patterns are updated by the current task, the model has a higher risk of forgetting. When $N_k=1$, only the pattern with the highest probability is updated, and tasks do not interfere with each other, so forgetting is less likely. Combining the analysis of the entire Fig.~\ref{fig:topk}, the effect of our proposed class-conditional routing is both intuitive and significant. It not only helps the model achieve strong learning ability for new tasks, comparable to the $N_k$ setting, but also provides excellent anti-forgetting performance, similar to $N_k=1$. This is due to the dynamic adjustment of $N_k$ based on uncertainty: for samples that are prone to misclassification, class-conditional routing assigns a larger $N_k$, enabling more extensive learning. For easier-to-classify samples, the model's learning ability is higher, and class-conditional routing assigns a smaller $N_k$ to enhance anti-forgetting.

\vspace{-4pt}
\paragraph{Sensitivity Analysis of Hyper-parameters $\alpha$ and $\beta$.} $\alpha$ and $\beta$ are the weights for $\mathcal{L}_{\text{Diff}}$ and $\mathcal{L}_{\text{Cont}}$, respectively. We conduct experiments on the sensitivity of them, and the results are shown in Table~\ref{tab:hyper}. It can be observed that when the weights are low, the optimization effect of the loss is relatively weak, both accuracy and forgetting rate do not perform well. Gradually increasing the weights, we find that after a certain point, the impact of the weights on the results becomes relatively stable, with no more drastic changes.

\begin{table}
    \centering
    \caption{Impact of hyper-parameters $\alpha$ and $\beta$ on performance in the CIFAR-100 dataset ($M=2k$). $\alpha$ and $\beta$ are the weights for $\mathcal{L}_{\text{Diff}}$ and $\mathcal{L}_{\text{Cont}}$, respectively. Acc. and AF represent the average accuracy and the average forgetting over 5 runs, respectively.}
    \vspace{-4pt}
    \resizebox{1\linewidth}{!}{
    \begin{tabular}{cccccc}
        \toprule
         \multicolumn{6}{c}{\textbf{{Impact of hyper-parameter $\alpha$}}}\\ \midrule 
          {\textbf{Value}} & 0.01& 0.1&  0.5&  1&  5\\ \midrule 
          {\textbf{Acc.}}  &  36.80{\scriptsize ±1.17}& 37.09{\scriptsize ±1.09}& 37.55{\scriptsize ±1.08} &  37.47{\scriptsize ±0.94} &37.57{\scriptsize ±0.81}\\ \midrule 
         {\textbf{AF}}  &  10.49{\scriptsize ±1.01}& 9.65{\scriptsize ±0.39}& 9.05{\scriptsize ±1.25} &  9.16{\scriptsize ±2.11} & 9.81{\scriptsize ±1.32}\\
         \midrule \midrule 
         \multicolumn{6}{c}{\textbf{{Impact of hyper-parameter $\beta$}}}\\ \midrule 
        {\textbf{Value}} & 0.1& 1&  5&  10&  25\\ \midrule 
         {\textbf{Acc.}}  &  36.50{\scriptsize ±1.24}& 37.30{\scriptsize ±1.03}& 37.40{\scriptsize ±0.33} &  37.26{\scriptsize ±0.61} &37.57{\scriptsize ±0.81}\\ \midrule 
         {\textbf{AF}}  &  10.24{\scriptsize ±0.43}& 10.12{\scriptsize ±1.09}& 10.10{\scriptsize ±0.60} & 9.98{\scriptsize ±0.66} & 9.81{\scriptsize ±1.32}\\
         
         \bottomrule
    \end{tabular}
    }
    \label{tab:hyper}
    \vspace{-13pt}
\end{table}

\begin{figure}
\begin{center}
    \subfloat[New-Task Average Accuracy w/o class-conditional routing]{
       \includegraphics[width=0.48\linewidth]{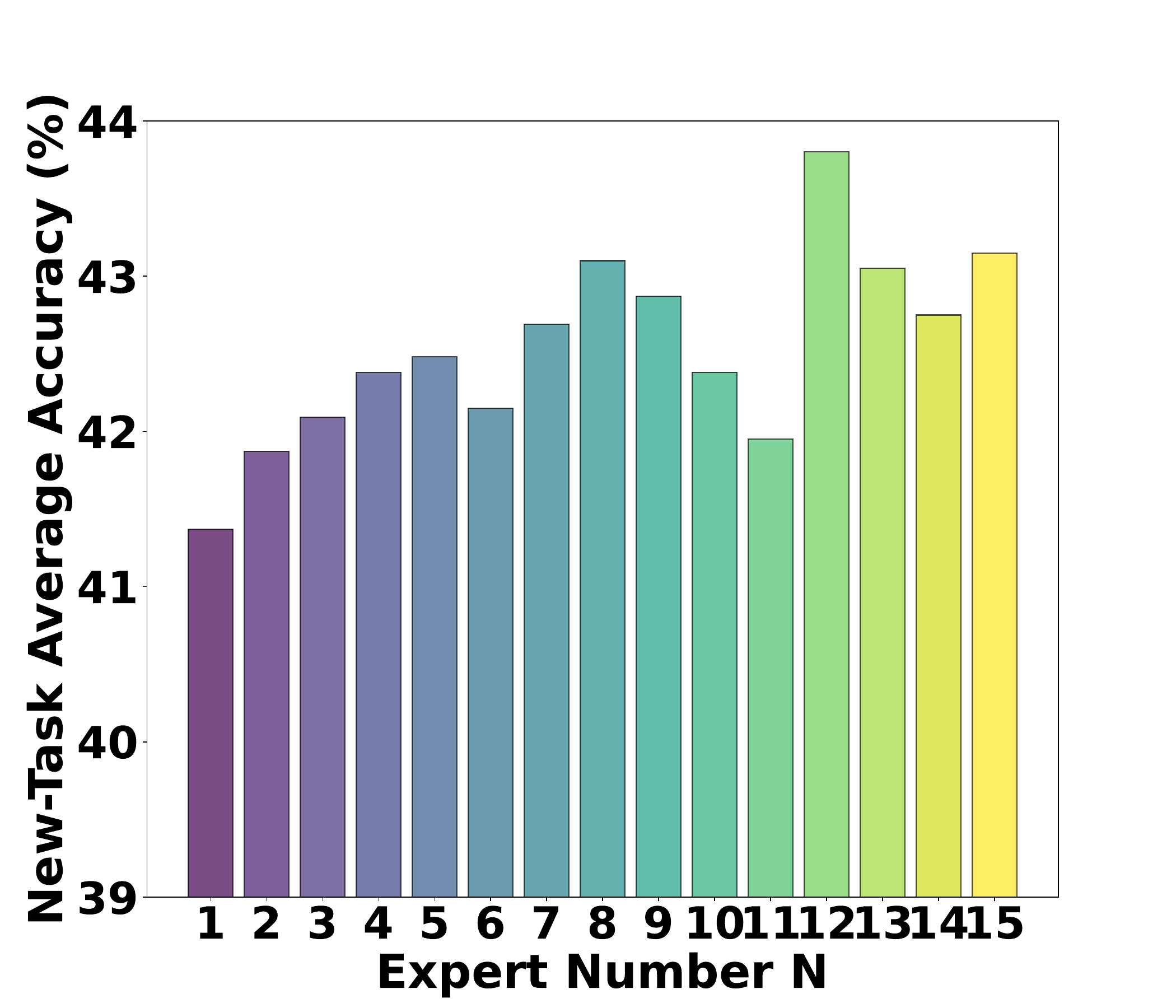}}
    \label{1a}
    \vspace{-2pt}
    \subfloat[Average Forgetting w/o class-conditional routing]{
       \includegraphics[width=0.485\linewidth]{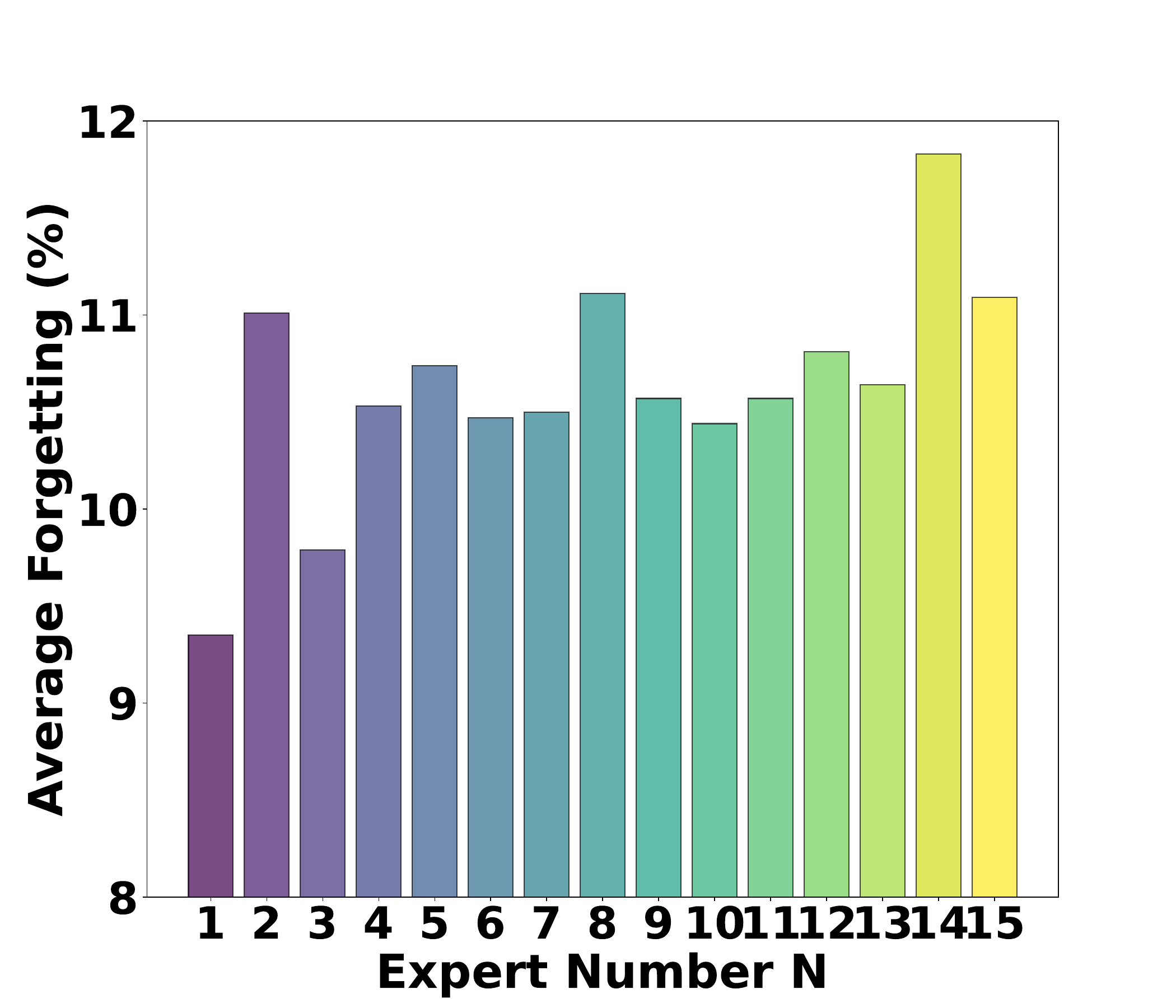}}
    \label{1a}
    \vspace{-2pt}
    \subfloat[New-Task Average Accuracy]{
       \includegraphics[width=0.48\linewidth]{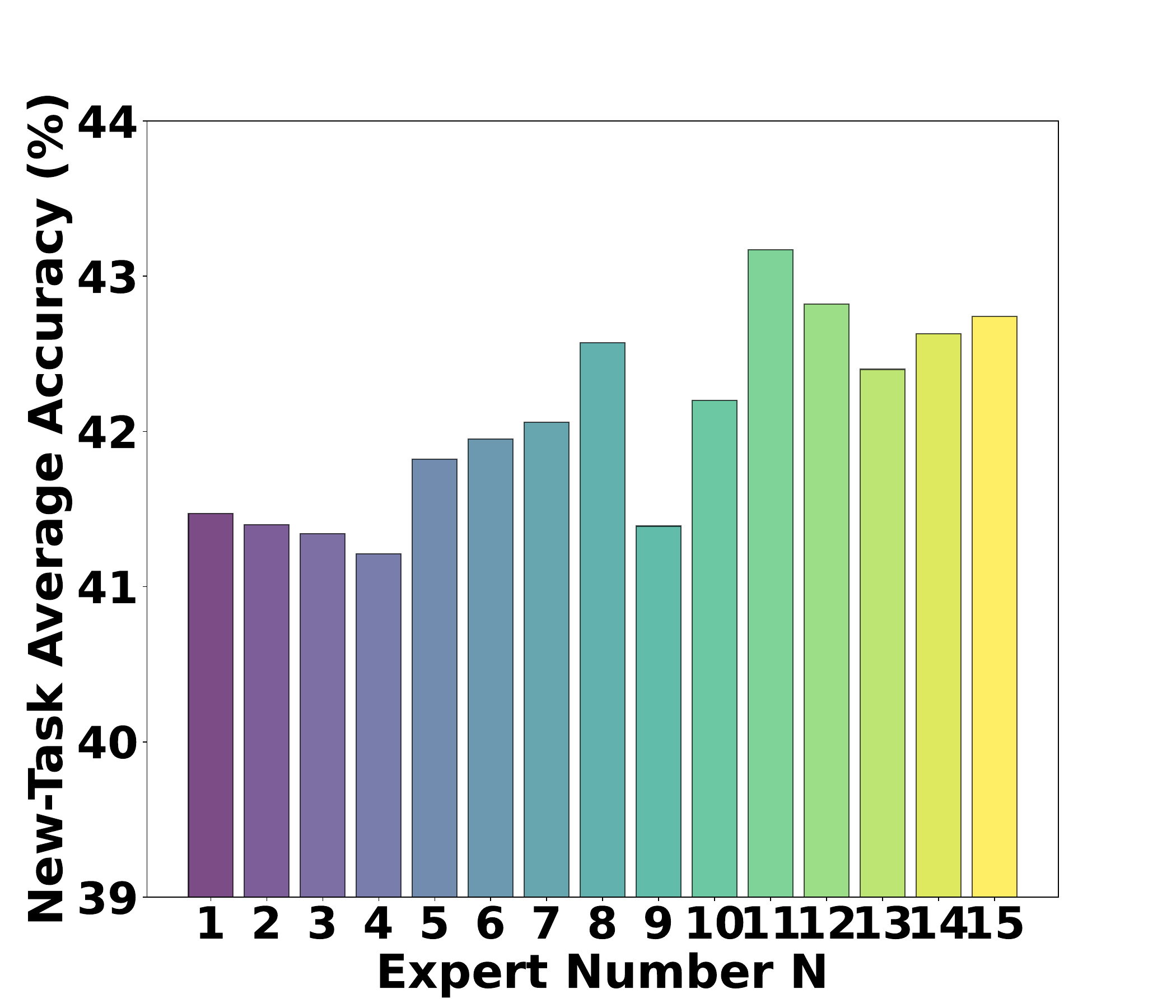}}
    \label{1a}
    \subfloat[Average Forgetting]{
        \includegraphics[width=0.485\linewidth]{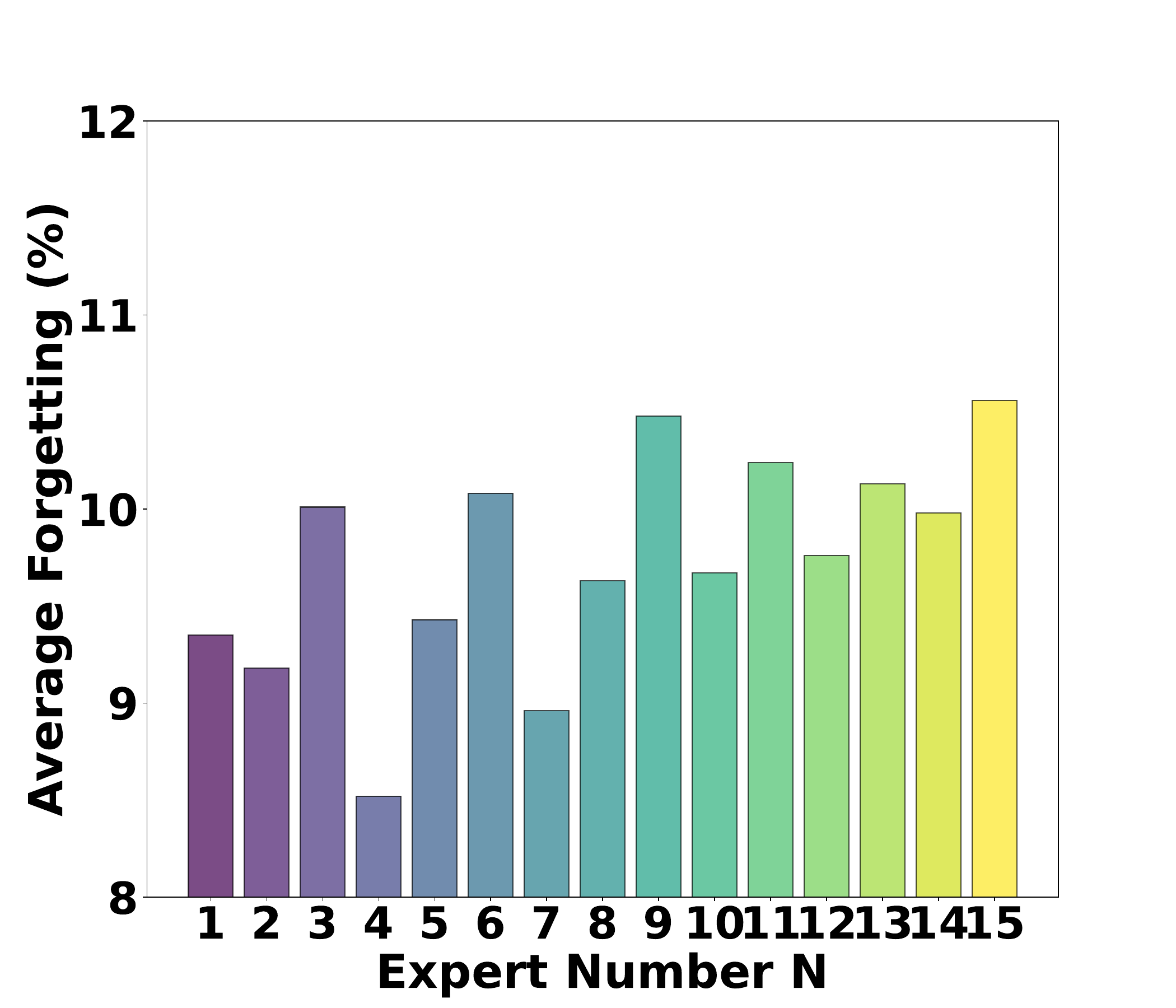}}
    \label{1b}
    \vspace{-4pt}
    \caption{Impact of pattern number $N$. The dataset used is CIFAR-100 ($M=2k$).}
    \label{fig:N} 
    \vspace{-22pt}
\end{center}
\end{figure}

\vspace{-4pt}
\paragraph{Sensitivity Analysis of total discretization patterns number $N$.} We conduct experiments on the total number of discretization patterns in the module. The results are shown in Fig.~\ref{fig:N}. As the number $N$ increases, the model gains a stronger ability to learn new tasks, but at the same time, its tendency to forget becomes more pronounced. Furthermore, by comparing (a)(b) and (c)(d) in Fig.~\ref{fig:N}, we can conclude that class-conditional routing can significantly reduce forgetting with only a slight decrease in new task accuracy, which in turn leads to an overall improvement in performance. This indirectly supports the effectiveness of class-conditional routing in our ablation studies.

\section{Conclusion}
In this paper, we propose a plug-and-play module S6MOD to enhance the adaptability of existing OCL methods by introducing selective state space models (SSMs) with a class-conditional mixture of discretization. 
By integrating a dynamic discretization mechanism and leveraging class-conditional strategies, our method can efficiently allocate discretization patterns based on class uncertainty, improving both the model's generalization and its ability to adapt to new data. Experimental results on multiple OCL datasets demonstrate that our method consistently optimize existing techniques, contributing to more robust OCL systems.
{
    \small
    \bibliographystyle{ieeenat_fullname}
    \bibliography{main}

\begin{thebibliography}{57}
\providecommand{\natexlab}[1]{#1}
\providecommand{\url}[1]{\texttt{#1}}
\expandafter\ifx\csname urlstyle\endcsname\relax
  \providecommand{\doi}[1]{doi: #1}\else
  \providecommand{\doi}{doi: \begingroup \urlstyle{rm}\Url}\fi

\bibitem[Abdulle et~al.(2012)Abdulle, Weinan, Engquist, and Vanden-Eijnden]{abdulle2012heterogeneous}
Assyr Abdulle, E Weinan, Bj{\"o}rn Engquist, and Eric Vanden-Eijnden.
\newblock The heterogeneous multiscale method.
\newblock \emph{Acta Numerica}, 21:\penalty0 1--87, 2012.

\bibitem[Aljundi et~al.(2018)Aljundi, Babiloni, Elhoseiny, Rohrbach, and Tuytelaars]{aljundi2018memory}
Rahaf Aljundi, Francesca Babiloni, Mohamed Elhoseiny, Marcus Rohrbach, and Tinne Tuytelaars.
\newblock Memory aware synapses: Learning what (not) to forget.
\newblock In \emph{ECCV}, pages 139--154, 2018.

\bibitem[Aljundi et~al.(2019)Aljundi, Belilovsky, Tuytelaars, Charlin, Caccia, Lin, and Page-Caccia]{aljundi2019online}
Rahaf Aljundi, Eugene Belilovsky, Tinne Tuytelaars, Laurent Charlin, Massimo Caccia, Min Lin, and Lucas Page-Caccia.
\newblock Online continual learning with maximal interfered retrieval.
\newblock \emph{NeurIPS}, 32, 2019.

\bibitem[Anthony et~al.(2024)Anthony, Tokpanov, Glorioso, and Millidge]{anthony2024blackmamba}
Quentin Anthony, Yury Tokpanov, Paolo Glorioso, and Beren Millidge.
\newblock Blackmamba: Mixture of experts for state-space models.
\newblock \emph{arXiv preprint arXiv:2402.01771}, 2024.

\bibitem[Buzzega et~al.(2020)Buzzega, Boschini, Porrello, Abati, and Calderara]{buzzega2020dark}
Pietro Buzzega, Matteo Boschini, Angelo Porrello, Davide Abati, and Simone Calderara.
\newblock Dark experience for general continual learning: a strong, simple baseline.
\newblock \emph{NeurIPS}, 33:\penalty0 15920--15930, 2020.

\bibitem[Chaudhry et~al.(2018{\natexlab{a}})Chaudhry, Dokania, Ajanthan, and Torr]{chaudhry2018riemannian}
Arslan Chaudhry, Puneet~K Dokania, Thalaiyasingam Ajanthan, and Philip~HS Torr.
\newblock Riemannian walk for incremental learning: Understanding forgetting and intransigence.
\newblock In \emph{ECCV}, pages 532--547, 2018{\natexlab{a}}.

\bibitem[Chaudhry et~al.(2018{\natexlab{b}})Chaudhry, Ranzato, Rohrbach, and Elhoseiny]{chaudhry2018efficient}
Arslan Chaudhry, Marc'Aurelio Ranzato, Marcus Rohrbach, and Mohamed Elhoseiny.
\newblock Efficient lifelong learning with a-gem.
\newblock \emph{arXiv preprint arXiv:1812.00420}, 2018{\natexlab{b}}.

\bibitem[Das et~al.(2023)Das, Kong, Sen, and Zhou]{das2023decoder}
Abhimanyu Das, Weihao Kong, Rajat Sen, and Yichen Zhou.
\newblock A decoder-only foundation model for time-series forecasting.
\newblock \emph{arXiv preprint arXiv:2310.10688}, 2023.

\bibitem[De~Lange et~al.(2021)De~Lange, Aljundi, Masana, Parisot, Jia, Leonardis, Slabaugh, and Tuytelaars]{de2021continual}
Matthias De~Lange, Rahaf Aljundi, Marc Masana, Sarah Parisot, Xu Jia, Ale{\v{s}} Leonardis, Gregory Slabaugh, and Tinne Tuytelaars.
\newblock A continual learning survey: Defying forgetting in classification tasks.
\newblock \emph{PAMI}, 44\penalty0 (7):\penalty0 3366--3385, 2021.

\bibitem[Eigen et~al.(2013)Eigen, Ranzato, and Sutskever]{eigen2013learning}
David Eigen, Marc'Aurelio Ranzato, and Ilya Sutskever.
\newblock Learning factored representations in a deep mixture of experts.
\newblock \emph{arXiv preprint arXiv:1312.4314}, 2013.

\bibitem[Fedus et~al.(2022)Fedus, Zoph, and Shazeer]{fedus2022switch}
William Fedus, Barret Zoph, and Noam Shazeer.
\newblock Switch transformers: Scaling to trillion parameter models with simple and efficient sparsity.
\newblock \emph{Journal of Machine Learning Research}, 23\penalty0 (120):\penalty0 1--39, 2022.

\bibitem[Gao et~al.(2023)Gao, Shan, Zhang, and Zhou]{NEURIPS2023_d7b3cef7}
Qiang Gao, Xiaojun Shan, Yuchen Zhang, and Fan Zhou.
\newblock Enhancing knowledge transfer for task incremental learning with data-free subnetwork.
\newblock In \emph{Advances in Neural Information Processing Systems}, pages 68471--68484. Curran Associates, Inc., 2023.

\bibitem[Gu and Dao(2023)]{gu2023mamba}
Albert Gu and Tri Dao.
\newblock Mamba: Linear-time sequence modeling with selective state spaces.
\newblock \emph{arXiv preprint arXiv:2312.00752}, 2023.

\bibitem[Gu et~al.(2021)Gu, Goel, and R{\'e}]{gu2021efficiently}
Albert Gu, Karan Goel, and Christopher R{\'e}.
\newblock Efficiently modeling long sequences with structured state spaces.
\newblock \emph{arXiv preprint arXiv:2111.00396}, 2021.

\bibitem[Guckenheimer and Holmes(2013)]{guckenheimer2013nonlinear}
John Guckenheimer and Philip Holmes.
\newblock \emph{Nonlinear oscillations, dynamical systems, and bifurcations of vector fields}.
\newblock Springer Science \& Business Media, 2013.

\bibitem[Gunasekara et~al.(2023)Gunasekara, Pfahringer, Gomes, and Bifet]{gunasekara2023survey}
Nuwan Gunasekara, Bernhard Pfahringer, Heitor~Murilo Gomes, and Albert Bifet.
\newblock Survey on online streaming continual learning.
\newblock In \emph{IJCAI}, pages 6628--6637, 2023.

\bibitem[Guo et~al.(2022)Guo, Liu, and Zhao]{guo2022online}
Yiduo Guo, Bing Liu, and Dongyan Zhao.
\newblock Online continual learning through mutual information maximization.
\newblock In \emph{ICML}, 2022.

\bibitem[Guo et~al.(2023)Guo, Liu, and Zhao]{guo2023dealing}
Yiduo Guo, Bing Liu, and Dongyan Zhao.
\newblock Dealing with cross-task class discrimination in online continual learning.
\newblock In \emph{CVPR}, 2023.

\bibitem[Jacobs et~al.(1991)Jacobs, Jordan, Nowlan, and Hinton]{6797059}
Robert~A. Jacobs, Michael~I. Jordan, Steven~J. Nowlan, and Geoffrey~E. Hinton.
\newblock Adaptive mixtures of local experts.
\newblock \emph{Neural Computation}, 3\penalty0 (1):\penalty0 79--87, 1991.

\bibitem[Kirkpatrick et~al.(2017)Kirkpatrick, Pascanu, Rabinowitz, Veness, Desjardins, Rusu, Milan, Quan, Ramalho, Grabska-Barwinska, et~al.]{kirkpatrick2017overcoming}
James Kirkpatrick, Razvan Pascanu, Neil Rabinowitz, Joel Veness, Guillaume Desjardins, Andrei~A Rusu, Kieran Milan, John Quan, Tiago Ramalho, Agnieszka Grabska-Barwinska, et~al.
\newblock Overcoming catastrophic forgetting in neural networks.
\newblock \emph{Proceedings of the national academy of sciences}, 114\penalty0 (13):\penalty0 3521--3526, 2017.

\bibitem[Krizhevsky(2009)]{Krizhevsky2009LearningML}
Alex Krizhevsky.
\newblock Learning multiple layers of features from tiny images.
\newblock 2009.

\bibitem[Le and Yang(2015)]{Le2015TinyIV}
Ya Le and Xuan~S. Yang.
\newblock Tiny imagenet visual recognition challenge.
\newblock 2015.

\bibitem[Li et~al.(2024)Li, Yang, Wu, Ghanem, Nie, and Zhang]{li2024mamba}
Xiaojie Li, Yibo Yang, Jianlong Wu, Bernard Ghanem, Liqiang Nie, and Min Zhang.
\newblock Mamba-fscil: Dynamic adaptation with selective state space model for few-shot class-incremental learning.
\newblock \emph{arXiv preprint arXiv:2407.06136}, 2024.

\bibitem[Li and Hoiem(2017)]{li2017learning}
Zhizhong Li and Derek Hoiem.
\newblock Learning without forgetting.
\newblock \emph{PAMI}, 40\penalty0 (12):\penalty0 2935--2947, 2017.

\bibitem[Lieber et~al.(2024)Lieber, Lenz, Bata, Cohen, Osin, Dalmedigos, Safahi, Meirom, Belinkov, Shalev-Shwartz, et~al.]{lieber2024jamba}
Opher Lieber, Barak Lenz, Hofit Bata, Gal Cohen, Jhonathan Osin, Itay Dalmedigos, Erez Safahi, Shaked Meirom, Yonatan Belinkov, Shai Shalev-Shwartz, et~al.
\newblock Jamba: A hybrid transformer-mamba language model.
\newblock \emph{arXiv preprint arXiv:2403.19887}, 2024.

\bibitem[Liu et~al.(2024)Liu, Tian, Zhao, Yu, Xie, Wang, Ye, and Liu]{Liu2024VMambaVS}
Yue Liu, Yunjie Tian, Yuzhong Zhao, Hongtian Yu, Lingxi Xie, Yaowei Wang, Qixiang Ye, and Yunfan Liu.
\newblock Vmamba: Visual state space model.
\newblock \emph{ArXiv}, abs/2401.10166, 2024.

\bibitem[Lopez-Paz and Ranzato(2017)]{lopez2017gradient}
David Lopez-Paz and Marc'Aurelio Ranzato.
\newblock Gradient episodic memory for continual learning.
\newblock \emph{NeurIPS}, 30, 2017.

\bibitem[Mai et~al.(2022)Mai, Li, Jeong, Quispe, Kim, and Sanner]{MAI202228}
Zheda Mai, Ruiwen Li, Jihwan Jeong, David Quispe, Hyunwoo Kim, and Scott Sanner.
\newblock Online continual learning in image classification: An empirical survey.
\newblock \emph{Neurocomputing}, 469:\penalty0 28--51, 2022.

\bibitem[Mehta et~al.(2022)Mehta, Gupta, Cutkosky, and Neyshabur]{mehta2022long}
Harsh Mehta, Ankit Gupta, Ashok Cutkosky, and Behnam Neyshabur.
\newblock Long range language modeling via gated state spaces.
\newblock \emph{arXiv preprint arXiv:2206.13947}, 2022.

\bibitem[Parisi et~al.(2019)Parisi, Kemker, Part, Kanan, and Wermter]{parisi2019continual}
German~I Parisi, Ronald Kemker, Jose~L Part, Christopher Kanan, and Stefan Wermter.
\newblock Continual lifelong learning with neural networks: A review.
\newblock \emph{Neural networks}, 113:\penalty0 54--71, 2019.

\bibitem[Pi{\'o}ro et~al.(2024)Pi{\'o}ro, Ciebiera, Kr{\'o}l, Ludziejewski, Krutul, Krajewski, Antoniak, Mi{\l}o{\'s}, Cygan, and Jaszczur]{pioro2024moe}
Maciej Pi{\'o}ro, Kamil Ciebiera, Krystian Kr{\'o}l, Jan Ludziejewski, Micha{\l} Krutul, Jakub Krajewski, Szymon Antoniak, Piotr Mi{\l}o{\'s}, Marek Cygan, and Sebastian Jaszczur.
\newblock Moe-mamba: Efficient selective state space models with mixture of experts.
\newblock \emph{arXiv preprint arXiv:2401.04081}, 2024.

\bibitem[Rebuffi et~al.(2017)Rebuffi, Kolesnikov, Sperl, and Lampert]{rebuffi2017icarl}
Sylvestre-Alvise Rebuffi, Alexander Kolesnikov, Georg Sperl, and Christoph~H Lampert.
\newblock icarl: Incremental classifier and representation learning.
\newblock In \emph{CVPR}, pages 2001--2010, 2017.

\bibitem[Rolnick et~al.(2019)Rolnick, Ahuja, Schwarz, Lillicrap, and Wayne]{rolnick2019experience}
David Rolnick, Arun Ahuja, Jonathan Schwarz, Timothy Lillicrap, and Gregory Wayne.
\newblock Experience replay for continual learning.
\newblock \emph{NeurIPS}, 32, 2019.

\bibitem[Ruan and Xiang(2024)]{ruan2024vm}
Jiacheng Ruan and Suncheng Xiang.
\newblock Vm-unet: Vision mamba unet for medical image segmentation.
\newblock \emph{arXiv preprint arXiv:2402.02491}, 2024.

\bibitem[Rusu et~al.(2016)Rusu, Rabinowitz, Desjardins, Soyer, Kirkpatrick, Kavukcuoglu, Pascanu, and Hadsell]{rusu2016progressive}
Andrei~A Rusu, Neil~C Rabinowitz, Guillaume Desjardins, Hubert Soyer, James Kirkpatrick, Koray Kavukcuoglu, Razvan Pascanu, and Raia Hadsell.
\newblock Progressive neural networks.
\newblock \emph{arXiv preprint arXiv:1606.04671}, 2016.

\bibitem[Seo et~al.(2024)Seo, Koh, Jeung, Lee, Kim, Lee, Cho, Choi, Kim, and Choi]{seo2024learning}
Minhyuk Seo, Hyunseo Koh, Wonje Jeung, Minjae Lee, San Kim, Hankook Lee, Sungjun Cho, Sungik Choi, Hyunwoo Kim, and Jonghyun Choi.
\newblock Learning equi-angular representations for online continual learning.
\newblock In \emph{CVPR}, pages 23933--23942, 2024.

\bibitem[Serra et~al.(2018)Serra, Suris, Miron, and Karatzoglou]{serra2018overcoming}
Joan Serra, Didac Suris, Marius Miron, and Alexandros Karatzoglou.
\newblock Overcoming catastrophic forgetting with hard attention to the task.
\newblock In \emph{ICML}, pages 4548--4557. PMLR, 2018.

\bibitem[Shazeer et~al.(2017)Shazeer, Mirhoseini, Maziarz, Davis, Le, Hinton, and Dean]{shazeer2017outrageously}
Noam Shazeer, Azalia Mirhoseini, Krzysztof Maziarz, Andy Davis, Quoc Le, Geoffrey Hinton, and Jeff Dean.
\newblock Outrageously large neural networks: The sparsely-gated mixture-of-experts layer.
\newblock \emph{arXiv preprint arXiv:1701.06538}, 2017.

\bibitem[van~der Maaten and Hinton(2008)]{JMLR:v9:vandermaaten08a}
Laurens van~der Maaten and Geoffrey Hinton.
\newblock Visualizing data using t-sne.
\newblock \emph{Journal of Machine Learning Research}, 9\penalty0 (86):\penalty0 2579--2605, 2008.

\bibitem[Vaswani(2017)]{vaswani2017attention}
A Vaswani.
\newblock Attention is all you need.
\newblock \emph{NeurIPS}, 2017.

\bibitem[Wang et~al.(2021)Wang, Zhang, Jia, Li, Bao, Ma, Zhu, and Zhong]{wang2021afec}
Liyuan Wang, Mingtian Zhang, Zhongfan Jia, Qian Li, Chenglong Bao, Kaisheng Ma, Jun Zhu, and Yi Zhong.
\newblock Afec: Active forgetting of negative transfer in continual learning.
\newblock \emph{NeurIPS}, 34:\penalty0 22379--22391, 2021.

\bibitem[Wang et~al.(2024{\natexlab{a}})Wang, Zhang, Su, and Zhu]{wang2024comprehensive}
Liyuan Wang, Xingxing Zhang, Hang Su, and Jun Zhu.
\newblock A comprehensive survey of continual learning: theory, method and application.
\newblock \emph{PAMI}, 2024{\natexlab{a}}.

\bibitem[Wang et~al.(2024{\natexlab{b}})Wang, Michel, Xiao, and Yamasaki]{wang2024improving}
Maorong Wang, Nicolas Michel, Ling Xiao, and Toshihiko Yamasaki.
\newblock Improving plasticity in online continual learning via collaborative learning.
\newblock In \emph{CVPR}, pages 23460--23469, 2024{\natexlab{b}}.

\bibitem[Wei et~al.(2023{\natexlab{a}})Wei, Ye, Huang, Zhang, and Shan]{OnPro}
Yujie Wei, Jiaxin Ye, Zhizhong Huang, Junping Zhang, and Hongming Shan.
\newblock Online prototype learning for online continual learning.
\newblock In \emph{ICCV}, pages 18764--18774, 2023{\natexlab{a}}.

\bibitem[Wei et~al.(2023{\natexlab{b}})Wei, Ye, Huang, Zhang, and Shan]{wei2023online}
Yujie Wei, Jiaxin Ye, Zhizhong Huang, Junping Zhang, and Hongming Shan.
\newblock Online prototype learning for online continual learning.
\newblock In \emph{ICCV}, 2023{\natexlab{b}}.

\bibitem[Wortsman et~al.(2020)Wortsman, Ramanujan, Liu, Kembhavi, Rastegari, Yosinski, and Farhadi]{wortsman2020supermasks}
Mitchell Wortsman, Vivek Ramanujan, Rosanne Liu, Aniruddha Kembhavi, Mohammad Rastegari, Jason Yosinski, and Ali Farhadi.
\newblock Supermasks in superposition.
\newblock \emph{NeurIPS}, 33:\penalty0 15173--15184, 2020.

\bibitem[Xie et~al.(2023)Xie, Yang, Cai, and He]{XIE202360}
Liang Xie, Yibo Yang, Deng Cai, and Xiaofei He.
\newblock Neural collapse inspired attraction–repulsion-balanced loss for imbalanced learning.
\newblock \emph{Neurocomputing}, 527:\penalty0 60--70, 2023.

\bibitem[Xing et~al.(2024)Xing, Ye, Yang, Liu, and Zhu]{xing2024segmamba}
Zhaohu Xing, Tian Ye, Yijun Yang, Guang Liu, and Lei Zhu.
\newblock Segmamba: Long-range sequential modeling mamba for 3d medical image segmentation.
\newblock In \emph{International Conference on Medical Image Computing and Computer-Assisted Intervention}, pages 578--588. Springer, 2024.

\bibitem[Yan et~al.(2024)Yan, Wang, Ma, and Zhong]{yan2024orchestrate}
Hongwei Yan, Liyuan Wang, Kaisheng Ma, and Yi Zhong.
\newblock Orchestrate latent expertise: Advancing online continual learning with multi-level supervision and reverse self-distillation.
\newblock In \emph{CVPR}, pages 23670--23680, 2024.

\bibitem[Yang et~al.(2022)Yang, Chen, Li, Xie, Lin, and Tao]{NEURIPS2022_f7f5f501}
Yibo Yang, Shixiang Chen, Xiangtai Li, Liang Xie, Zhouchen Lin, and Dacheng Tao.
\newblock Inducing neural collapse in imbalanced learning: Do we really need a learnable classifier at the end of deep neural network?
\newblock In \emph{NeurIPS}, pages 37991--38002. Curran Associates, Inc., 2022.

\bibitem[Yang et~al.(2023)Yang, Yuan, Li, Lin, Torr, and Tao]{yang2023neural}
Yibo Yang, Haobo Yuan, Xiangtai Li, Zhouchen Lin, Philip Torr, and Dacheng Tao.
\newblock Neural collapse inspired feature-classifier alignment for few-shot class incremental learning.
\newblock \emph{arXiv preprint arXiv:2302.03004}, 2023.

\bibitem[Yoon et~al.(2017)Yoon, Yang, Lee, and Hwang]{yoon2017lifelong}
Jaehong Yoon, Eunho Yang, Jeongtae Lee, and Sung~Ju Hwang.
\newblock Lifelong learning with dynamically expandable networks.
\newblock \emph{arXiv preprint arXiv:1708.01547}, 2017.

\bibitem[Yuksel et~al.(2012)Yuksel, Wilson, and Gader]{6215056}
Seniha~Esen Yuksel, Joseph~N. Wilson, and Paul~D. Gader.
\newblock Twenty years of mixture of experts.
\newblock \emph{IEEE Transactions on Neural Networks and Learning Systems}, 23\penalty0 (8):\penalty0 1177--1193, 2012.

\bibitem[Zenke et~al.(2017)Zenke, Poole, and Ganguli]{zenke2017continual}
Friedemann Zenke, Ben Poole, and Surya Ganguli.
\newblock Continual learning through synaptic intelligence.
\newblock In \emph{ICML}, pages 3987--3995. PMLR, 2017.

\bibitem[Zhong et~al.(2023)Zhong, Cui, Yang, Wu, Qi, Zhang, and Jia]{Zhong_2023_CVPR}
Zhisheng Zhong, Jiequan Cui, Yibo Yang, Xiaoyang Wu, Xiaojuan Qi, Xiangyu Zhang, and Jiaya Jia.
\newblock Understanding imbalanced semantic segmentation through neural collapse.
\newblock In \emph{CVPR}, pages 19550--19560, 2023.

\bibitem[Zhu et~al.(2024)Zhu, Liao, Zhang, Wang, Liu, and Wang]{zhu2024vision}
Lianghui Zhu, Bencheng Liao, Qian Zhang, Xinlong Wang, Wenyu Liu, and Xinggang Wang.
\newblock Vision mamba: Efficient visual representation learning with bidirectional state space model.
\newblock \emph{arXiv preprint arXiv:2401.09417}, 2024.

\bibitem[Zoph et~al.(2022)Zoph, Bello, Kumar, Du, Huang, Dean, Shazeer, and Fedus]{zoph2022st}
Barret Zoph, Irwan Bello, Sameer Kumar, Nan Du, Yanping Huang, Jeff Dean, Noam Shazeer, and William Fedus.
\newblock St-moe: Designing stable and transferable sparse expert models.
\newblock \emph{arXiv preprint arXiv:2202.08906}, 2022.

\end{thebibliography}
}

\clearpage
\appendix
\setcounter{page}{1}
\maketitlesupplementary


\section{Implementation details}

\subsection{Training details}
In Table \ref{tab:hp}, we provide the hyper-parameter settings for our method when ER is used as the baseline. As shown in the table, for the same dataset, we tend to set the total number of patterns $N$ to a fixed value and set $\alpha$ and $\beta$ to a ratio of $1:5$. When we need to reduce the impact of our module, we can proportionally decrease the weights. Actually, different hyper-parameter settings help to unleash the potential of various classifiers (linear classifier, ETF classifier, NCM classifier). hyper-parameters not mentioned in the table remain consistent with the original baseline.
\begin{table*}
    \centering
    \resizebox{0.66\linewidth}{!}{
    \begin{tabular}{lccccccccc}
\toprule
\multicolumn{1}{c}{Dataset} & \multicolumn{2}{c}{CIFAR10} & \multicolumn{3}{c}{CIFAR100} & \multicolumn{3}{c}{Tiny-ImageNet}\\
\cmidrule(lr){1-1}
\cmidrule(lr){2-3}
\cmidrule(lr){4-6}
\cmidrule(lr){7-9}
\multicolumn{1}{c}{Memory Size $M$} & 500 & 1000 & 1000 & 2000 & 5000 & 2000 & 5000 & 10000\\
\midrule
$N$ & 10 & 10 & 8 & 8 & 8 & 10 & 10 & 10\\
$\alpha$ & 1 & 1 & 1 & 1 & 1 & 1 & 1 & 0.5\\
$\beta$ & 5 & 5 & 5 & 5 & 5 & 5 & 5 & 2.5\\

\bottomrule
\end{tabular}
    }
    \caption{The hyper-parameter settings for our S6MOD on ER.}
    \label{tab:hp}
\end{table*}

\subsection{Dataset}
As stated in Sec.~\ref{experiments} (Experiments), we primarily conduct experimental validation on three datasets: CIFAR10, CIFAR100, and TinyImageNet. It is important to note that the sample sizes and the number of classes vary across these datasets, which may lead to the use of different hyper-parameters in our method. Our experimental implementation follows the guidelines of CCLDC \cite{wang2024improving}. Specifically:
\paragraph{CIFAR-10} is a dataset composed of 10 classes, which we divide into 5 tasks, with each task containing 2 classes. It includes a total of 50,000 training samples and 10,000 test samples, with image dimensions of 32$\times$32.
\paragraph{CIFAR-100} consists of 100 classes, divided into 10 tasks, with each task containing 10 classes. It also contains 50,000 training samples and 10,000 test samples, with image dimensions of 32$\times$32.
\paragraph{TinyImageNet} comprises 200 classes, divided into 100 tasks, with each task containing 2 classes. It includes 100,000 training samples and 10,000 test samples, with image dimensions of 64$\times$64.

\subsection{Pseudo-code}
To facilitate understanding and usage of our proposed plug-and-play module, S6MOD, we provide pseudo-code in Algorithm \ref{code:pseudo_code} to demonstrate how to integrate S6MOD with the current baseline. For simplicity, we omit the workflows of $\mathcal{L}_{\text{DR}}$ and $\mathcal{L}_{\text{z}}$, as well as the samples in the memory buffer. 


\begin{spacing}{0.8}
\begin{algorithm}[t]
\scriptsize
\begin{minted}{python}
# model: the whole model
# model.logits: logit function of model (base classification)
# model.S6MOD: obtain features using S6MOD
# model.ETF: ETF logit function of model (ETF classification)
# cos_sim: cosine similarity calculation function
# optim: optimizer for model
for x, y in dataloader:
  # Baseline loss
  pred_base = model.logits(x)
  loss_base = criterion_baseline(model, x, y)

  # S6MOD loss
  fea, deltas = model.S6MOD(x)
  pred_etf = model.ETF(fea)

  loss_Diff = kl_div(pred_base, pred_etf)
  loss_Cont = 0
  for i in range(len(y)):
    for j in range(i+1, len(y)):
        if y[i]==y[j]:
            loss_Cont -= cos_sim(deltas[i], deltas[j])
        else:
            loss_Cont += cos_sim(deltas[i], deltas[j])
    

  # hyperparameters alpha and beta
  loss_S6MOD = loss_DR + alpha*loss_Diff + beta*loss_Cont + loss_z 
  loss = loss_base + loss_S6MOD
  
  optim.zero_grad()
  loss.backward()
  optim.step()
\end{minted}
\vspace{-4pt}
\caption{PyTorch-like pseudo-code of S6MOD to integrate to other baselines.}
\label{code:pseudo_code}
\end{algorithm}
\end{spacing}

\subsection{Metrics}
We use three commonly employed evaluation metrics Average Accuracy (Acc), Average Forgetting (AF) and New-Task Average Accuracy (N-Acc) in the main text \cite{wang2024improving}, and we will introduce their definitions in detail here.

In continual learning, after each task \( t \) is completed, the model needs to be tested on all previously learned tasks \( \{1, 2, \dots, t\} \). The Acc is defined as:

\begin{equation}
Acc_T = \frac{1}{T} \sum_{t=1}^T A_{t,T},
\end{equation}
where \( T \) is the total number of tasks, and \( A_{t,T} \) is the test accuracy on task \( t \) after learning task \( T \):

\begin{equation}
A_{t,T} = \frac{\sum_{i=1}^{N_t} 1(\hat{y}_{i,t} = y_{i,t})}{N_t}.
\end{equation}
Here, \( N_t \) is the number of samples in task \( t \), \( \hat{y}_{i,t} \) is the predicted class of the \( i \)-th sample, and \( y_{i,t} \) is the true class of the \( i \)-th sample.

The Average Forgetting (AF) is the average of the forgetting rates over all tasks. It provides an overall measure of how much the model forgets across all previously learned tasks as new tasks are added. A low AF indicates that the model effectively retains knowledge from previous tasks, while a high AF suggests that the model suffers from significant forgetting when learning new tasks. AF is defined as:
\begin{equation}
    AF = \frac{1}{T-1} \sum_{t=2}^T FR_t,
\end{equation}
where $T$ is the total number of tasks. \( FR_t \) is the Forgetting Rate for task \( t \), defined as:
\begin{equation}
    FR_t = \max_{i \in \{1, \dots, t-1\}} \left( A_{i,i} - A_{i,t} \right),
\end{equation}
where \( A_{i,i} \) is the accuracy on task \( i \) after learning task \( i \), and \( A_{i,t} \) is the accuracy on task \( i \) after learning task \( t \). The AF is averaged over all tasks after the first one, as the first task does not cause any forgetting.

The New-Task Average Accuracy (N-Acc) is the average accuracy of the model on all tasks when they are first learned. This metric provides an overall measure of how well the model performs on each task at the time it is introduced, without considering any changes in performance as other tasks are learned later. N-Acc is defined as:

\begin{equation}
    \text{N-Acc} = \frac{1}{T} \sum_{t=1}^{T} A_{t,t},
\end{equation}
where \( T \) is the total number of tasks and \( A_{t,t} \) is the accuracy on task \( t \) immediately after task \( t \) is learned, \emph{i.e.}, when the model first encounters the task. This metric directly reflects the model's ability to learn new tasks.

\section{Extra Experiments}



\subsection{Performance with NCM classifier.}
The Nearest Class Mean (NCM) classifier is a simple yet effective classification method, often used as a component in continual learning scenarios.
To further demonstrate that our method also learns more generalizable and discriminative features with NCM, we use an NCM classifier to test our method. As shown in Table \ref{tab:ncm} and Table \ref{tab:ncm2}, our method achieves superior performance when using the NCM classifier. This indicates that our method is also compatible with NCM classifier to learn more discriminative features.


\begin{table}
    \centering
    \resizebox{0.85\linewidth}{!}{
    \begin{tabular}{lcc}
\toprule
\multicolumn{1}{c}{Method} & NCM Acc. ↑ & Logit Acc. ↑\\
\midrule
ER & 64.31{\scriptsize ±0.98}& 62.32{\scriptsize ±4.13}\\
ER + Ours & 67.24{\scriptsize ±2.32} & 65.80{\scriptsize ±2.16} \\
\midrule
OCM & 72.47{\scriptsize ±1.04} & 73.15{\scriptsize ±1.05}\\
OCM + Ours & 76.36{\scriptsize ±0.66} & 75.31{\scriptsize ±1.10}\\
\midrule
OCM-CCLDC & 74.80{\scriptsize ±1.72}& 77.66{\scriptsize ±1.46}\\
OCM-CCLDC + Ours & 79.37{\scriptsize ±0.89} & 78.21{\scriptsize ±1.03}\\
\bottomrule
\end{tabular}
    }
    \caption{Final average accuracy on CIFAR-10 ($M=1k$), with and without our method on NCM and Logit predictions.}
    \label{tab:ncm}
\end{table}

\begin{table}
    \centering
    \resizebox{0.85\linewidth}{!}{
    \begin{tabular}{lcc}
\toprule
\multicolumn{1}{c}{Method} & NCM Acc. ↑ & Logit Acc. ↑\\
\midrule
ER & 36.40{\scriptsize ±0.81}& 31.89{\scriptsize ±1.45}\\
ER + Ours & 36.99{\scriptsize ±0.65} & 34.55{\scriptsize ±1.66} \\
\midrule
OCM & 37.76{\scriptsize ±0.70} & 35.69{\scriptsize ±1.36}\\
OCM + Ours & 39.98{\scriptsize ±1.19} & 38.97{\scriptsize ±2.28}\\
\midrule
OCM-CCLDC & 40.28{\scriptsize ±1.08}& 43.34{\scriptsize ±1.51}\\
OCM-CCLDC + Ours & 44.55{\scriptsize ±1.42} & 44.40{\scriptsize ±2.26}\\
\bottomrule
\end{tabular}
    }
    \caption{Final average accuracy on CIFAR-100 ($M=2k$), with and without our method on NCM and Logit predictions..}
    \label{tab:ncm2}
\end{table}


\subsection{More T-SNE visualization.} 
\begin{figure*}
\begin{center}
    \subfloat[ER]{
       \includegraphics[width=0.24\linewidth]{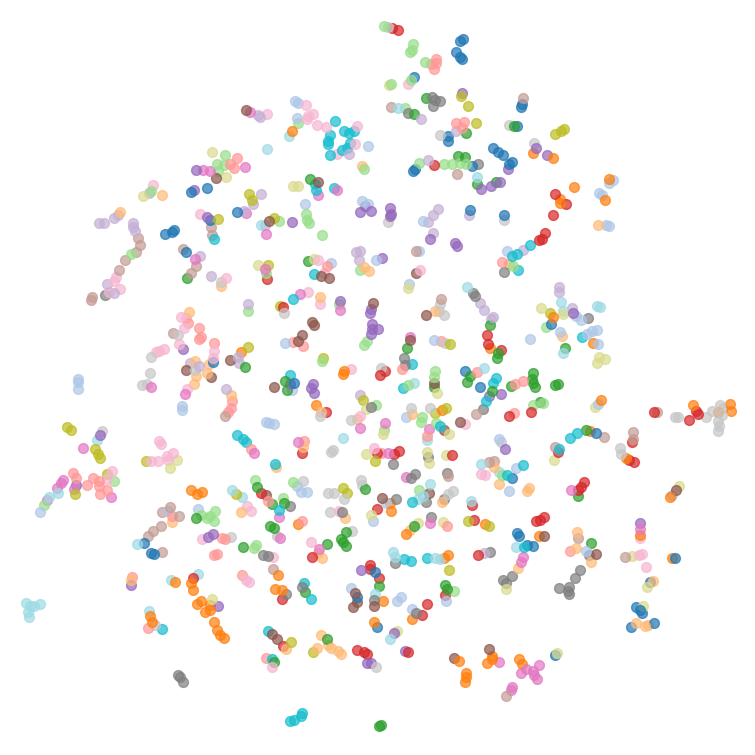}}
    \label{1a}
    \subfloat[ER + Ours]{
        \includegraphics[width=0.24\linewidth]{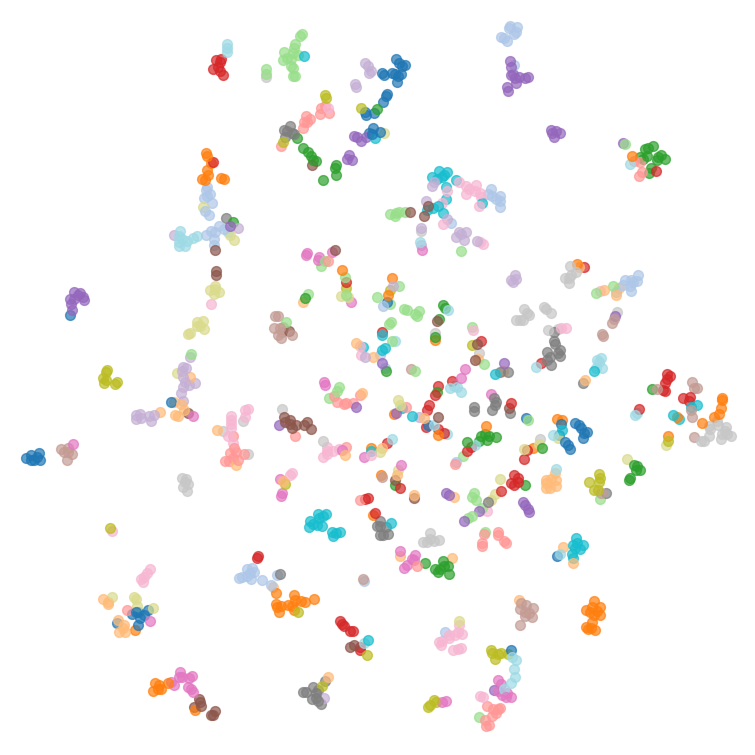}}
    \label{1b}
        \subfloat[OCM]{
       \includegraphics[width=0.24\linewidth]{imgs/embedding_plot_ocm_c100.pdf}}
    \label{2a}
    \subfloat[OCM + Ours]{
        \includegraphics[width=0.24\linewidth]{imgs/embedding_plot_ocm_mod_c100.pdf}}
    \label{2b}\\
    \subfloat[OnPro]{
       \includegraphics[width=0.24\linewidth]{imgs/embedding_plot_onpro_c100m2000.pdf}}
    \label{3a}
    \subfloat[OnPro + Ours]{
        \includegraphics[width=0.24\linewidth]{imgs/embedding_plot_onpro_mod_c100m2000.pdf}}
    \label{3b}
    \subfloat[OCM-CCLDC]{
       \includegraphics[width=0.24\linewidth]{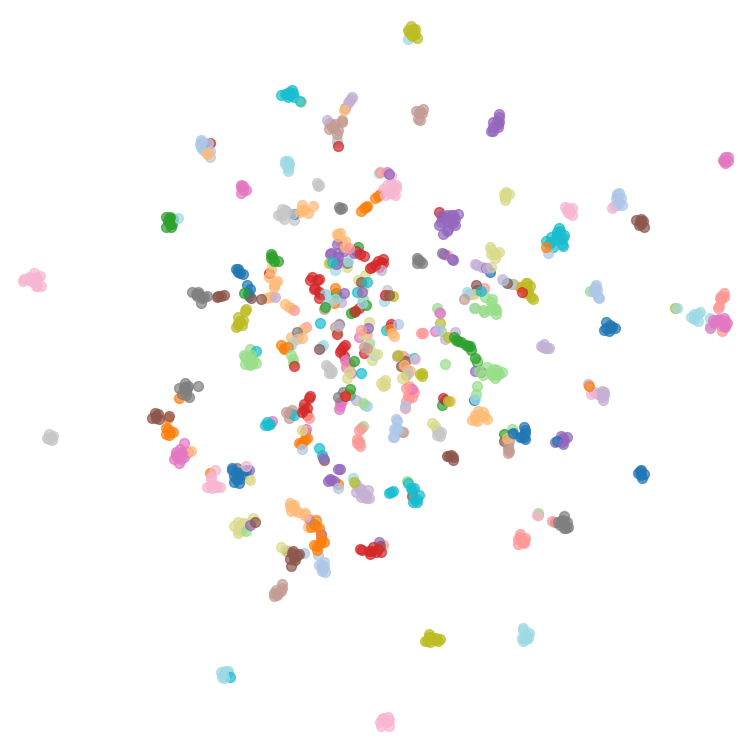}}
    \label{4a}
    \subfloat[OCM-CCLDC + Ours]{
        \includegraphics[width=0.24\linewidth]{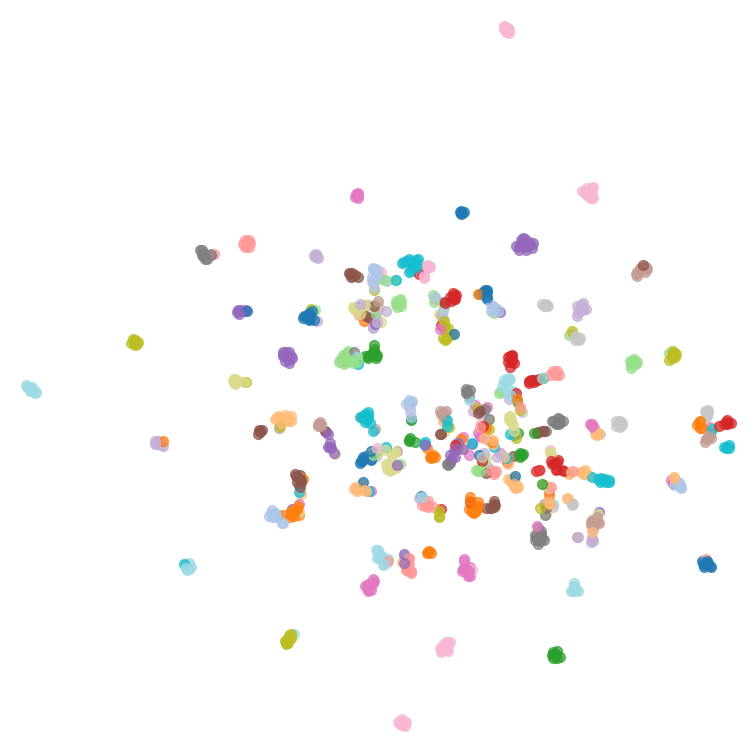}}
    \label{4b} \\
    \subfloat[OnPro-CCLDC]{
       \includegraphics[width=0.24\linewidth]{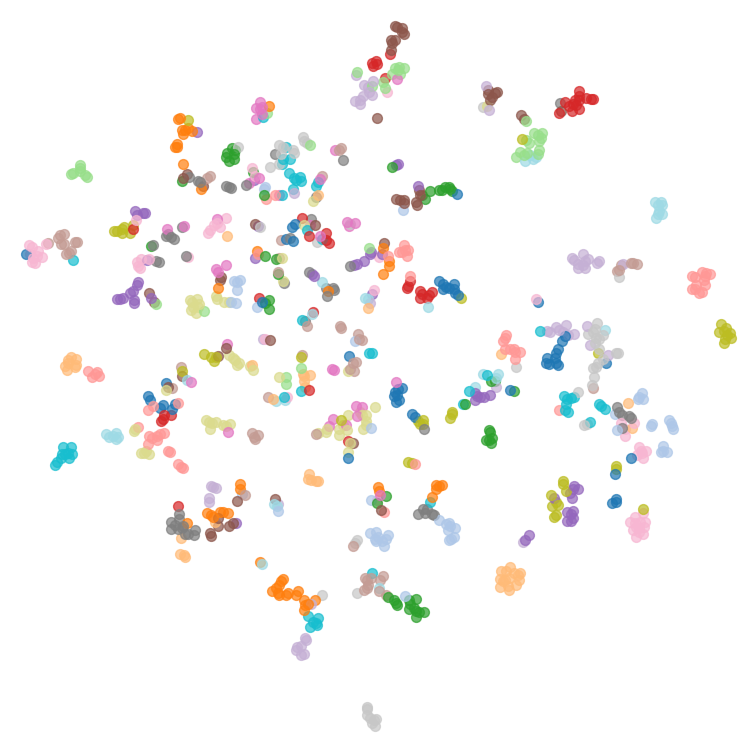}}
    \label{3a}
    \subfloat[OnPro-CCLDC + Ours]{
        \includegraphics[width=0.24\linewidth]{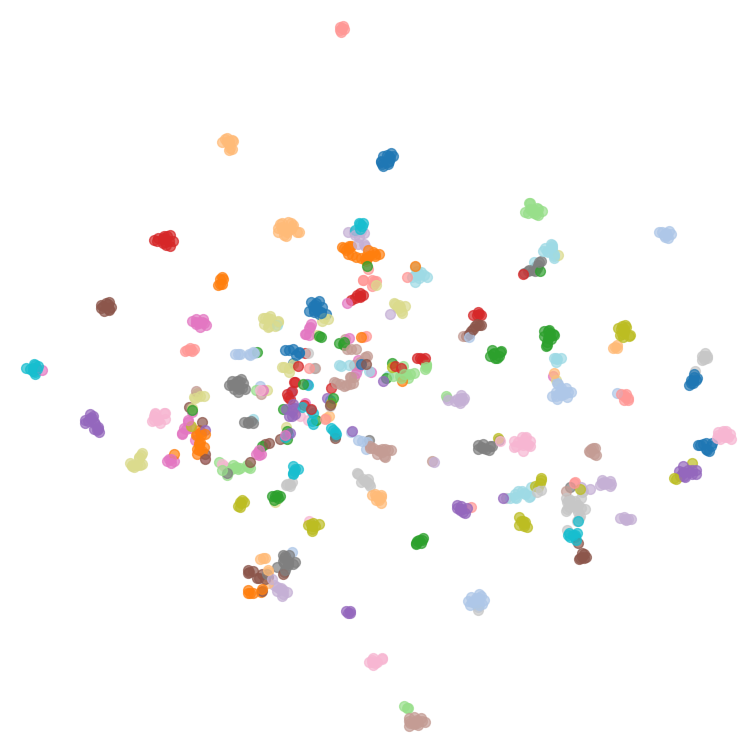}}
    \label{3b}
    \subfloat[MOSE]{
       \includegraphics[width=0.24\linewidth]{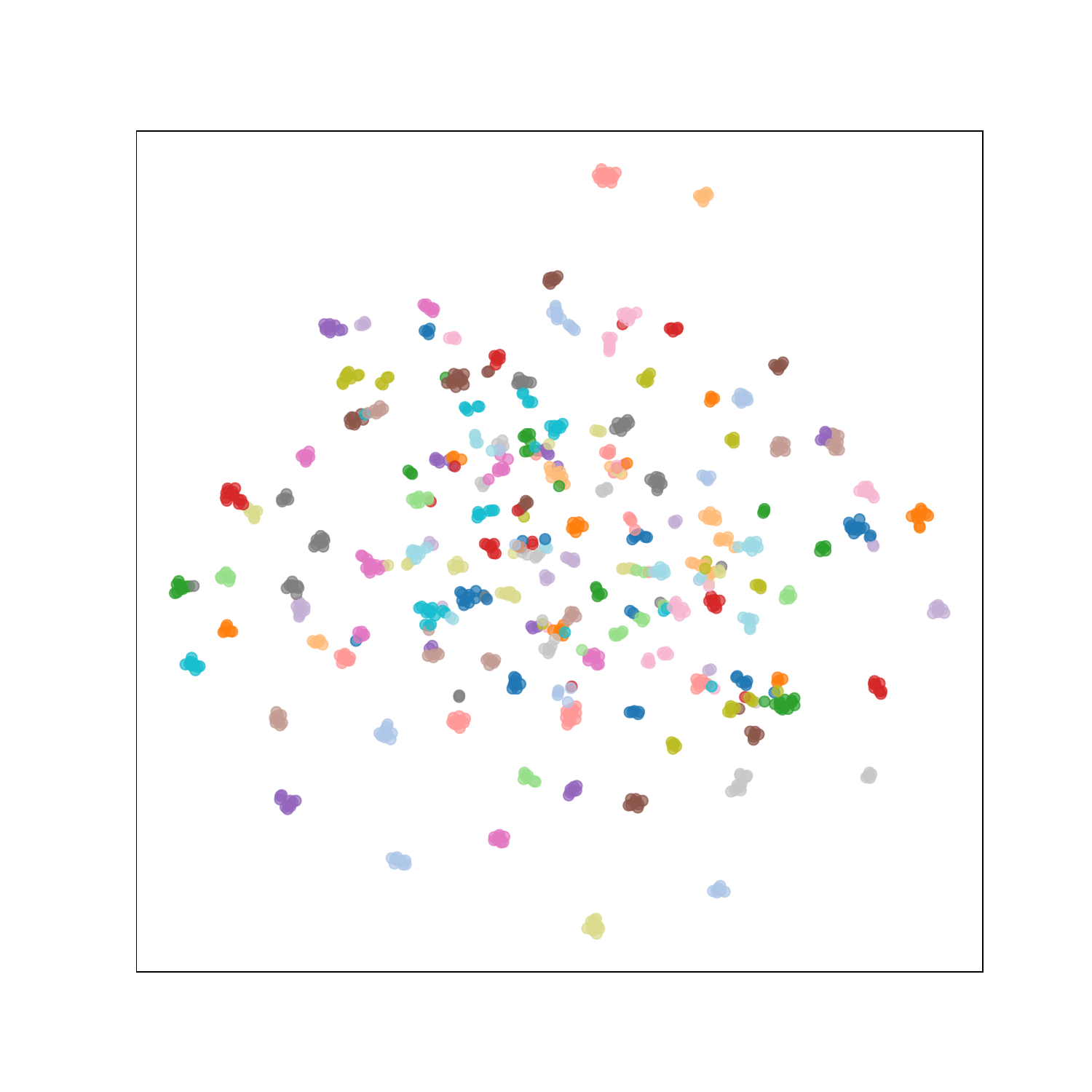}}
    \label{4a}
    \subfloat[MOSE + Ours]{
        \includegraphics[width=0.24\linewidth]{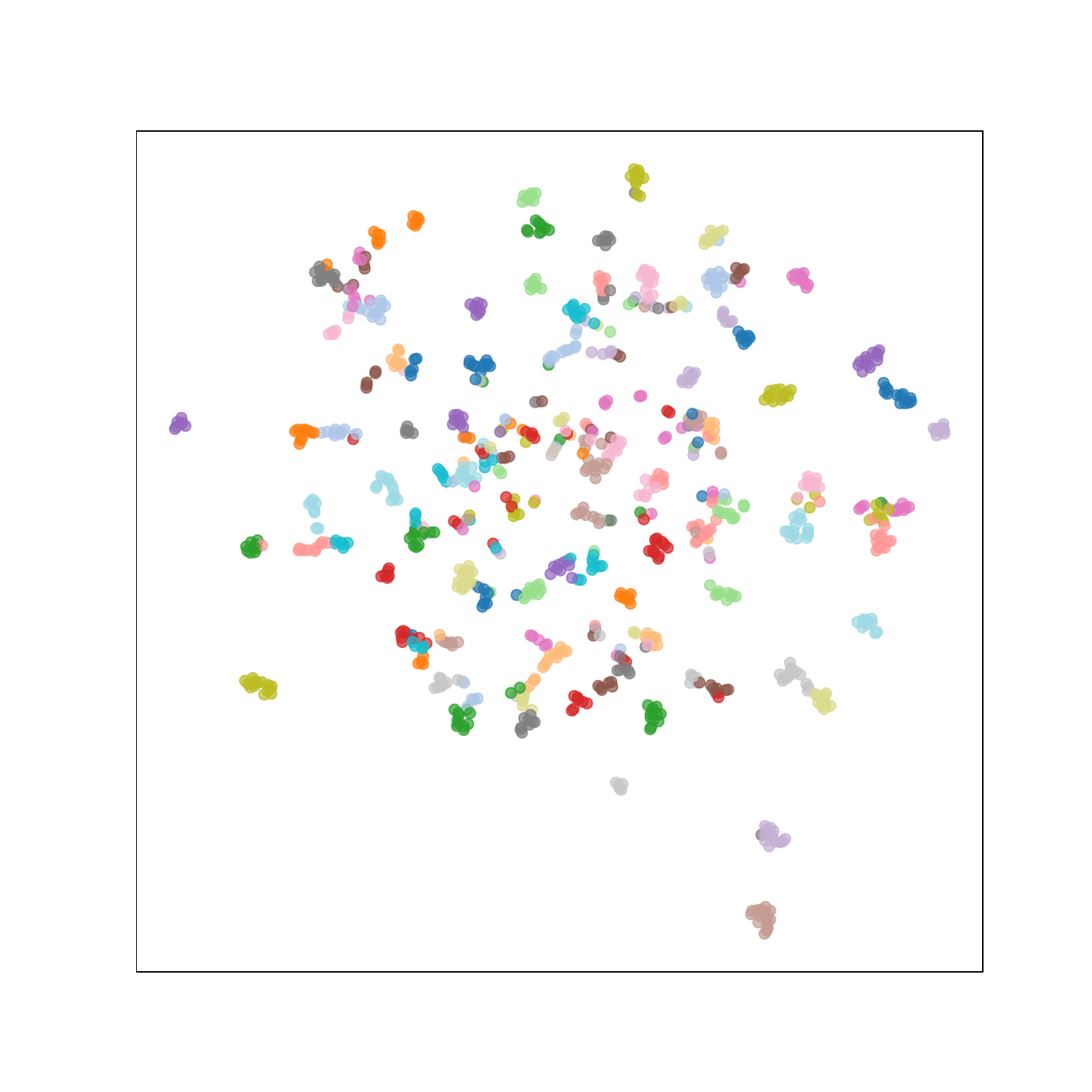}}
    \label{4b} \\
    \caption{T-SNE visualization of features before classification of memory data at the end of training on CIFAR-100 ($M=2k$).}
    \label{fig:t-SNE_extra} 
\end{center}
\end{figure*}

As described in Sec.~\ref{sec:intro} (Introduction) and demonstrated in ``Analysis of Feature Embedding," incorporating S6MOD helps the model learn more generalizable and discriminative features. To further validate this, we present comprehensive t-SNE visualizations in Fig.~\ref{fig:t-SNE_extra}, explicitly showcasing the superiority of our method on more baseline methods. Given that the MOSE and MOE-MOSE structures are identical, with the only difference being during inference, we only report the features of MOSE here.

\end{document}